\documentclass[sigconf]{acmart}
\usepackage{multirow}
\usepackage{tabularx}
\usepackage{makecell}
\usepackage{algorithmic}
\usepackage{algorithm}
\usepackage{bm}
\usepackage{booktabs}
\usepackage{amsmath}
\usepackage{graphicx}
\usepackage{caption}
\usepackage{utfsym}
\usepackage{diagbox}
\usepackage{subcaption}
\usepackage[table]{xcolor}
\AtBeginDocument{%
  }

\setcopyright{acmlicensed}
\copyrightyear{2026}
\acmYear{2026}
\setcopyright{cc}
\setcctype{by}
\acmDOI{XXXXXXX.XXXXXXX}
\acmConference[MM '26] {Proceedings of the 34th ACM International Conference on Multimedia}{November 10--14, 2026}{Rio de Janeiro, Brazil.}
\acmBooktitle{Proceedings of the 34th ACM International Conference on Multimedia (MM '26), November 10--14, 2026, Rio de Janeiro, Brazil}
\acmISBN{978-1-4503-XXXX-X/2018/06}

\begin{document}

\title{IDATA: Scalable Invertible Diffusion for Unrestricted Adversarial Transfer Attack}


\author{Yi Pan}
\orcid{0009-0004-4924-7562}
\affiliation{%
  \department{College of Computer Science and Technology}
  \institution{National University of Defense Technology}
  \city{Changsha}
  \country{China}}
\email{panyi_jsjy@nudt.edu.cn}

\author{Jun-Jie Huang}
\authornote{Corresponding author.}
\orcid{0000-0003-2986-4665}
\affiliation{%
  \department{College of Computer Science and Technology}
  \institution{National University of Defense Technology}
  \city{Changsha}
  \country{China}
}
\email{jjhuang@nudt.edu.cn}

\author{Tianrui Liu}
\orcid{0000-0001-7926-3310}
\affiliation{%
  \department{College of Computer Science and Technology}
  \institution{National University of Defense Technology}
  \city{Changsha}
  \country{China}
}
\email{trliu@nudt.edu.cn}

\author{Zihan Chen}
\orcid{0009-0007-8387-5687}
\affiliation{%
  \department{College of Computer Science and Technology}
  \institution{National University of Defense Technology}
  \city{Changsha}
  \country{China}
}
\email{chenzihan21@nudt.edu.cn}

\author{Lin Liu}
\orcid{0000-0002-5930-8881}
\affiliation{%
  \department{College of Computer Science and Technology}
  \institution{National University of Defense Technology}
  \city{Changsha}
  \country{China}
}
\email{liulin16@nudt.edu.cn}

\author{Zhao Wentao}
\authornote{Corresponding author.}
\orcid{0000-0002-2906-3238}
\affiliation{%
  \department{College of Computer Science and Technology}
  \institution{National University of Defense Technology}
  \city{Changsha}
  \country{China}
}
\email{wtzhao@nudt.edu.cn}

\renewcommand{\shortauthors}{Yi Pan et al.}

\begin{abstract}
Unrestricted adversarial transfer attacks are important for evaluating the black-box robustness of deep visual models. Diffusion-based attacks have shown promising transferability and visual imperceptibility by optimizing adversarial perturbations along denoising trajectories in latent space. However, existing methods are limited by two challenges: memory-intensive multistep backpropagation and frequency-agnostic perturbation over intermediate latents. To address these issues, we propose \textbf{IDATA}, a memory-efficient diffusion framework for unrestricted adversarial transfer attack. IDATA consists of two key components: an Invertible Diffusion Module (IDM) and a Low-Frequency Constraint Module (LFCM). Specifically, IDM reformulates adversarial optimization over diffusion trajectories as an invertible process, enabling constant-memory backpropagation through on-demand reconstruction of intermediate states instead of storing the full denoising chain. Moreover, LFCM leverages Discrete Wavelet Transform (DWT) to decompose latent variables into low- and high-frequency components, restricting perturbations to semantically stable low-frequency subspaces, thereby improving transferability while preserving visual imperceptibility. Extensive experiments on multiple benchmarks and diverse model architectures demonstrate that IDATA consistently outperforms state-of-the-art baselines in attack success rate, memory efficiency, and visual imperceptibility. These results suggest that IDATA is a promising tool for black-box robustness evaluation of deep visual models. Code is available at \url{https://github.com/colourful-pan/IDATA}.
\end{abstract}




\maketitle

\section{Introduction}

Deep visual models are now widely deployed in modern multimedia systems, supporting safety-critical applications such as video surveillance~\cite{liu2025personalized,bao2022hierarchical}, biometric authentication~\cite{mi2022duetface,li2023err}, and autonomous driving~\cite{ye2025safedriverag,han2022physical}. Their widespread deployment has made adversarial robustness a central challenge in multimedia security, since visually inconspicuous adversarial examples can still induce severe model failures in real-world pipelines~\cite{moosavi2017universal,chen2022adversarial,wang2023multi,huang2024dempaa}.

Among existing attack paradigms, adversarial transfer attacks are particularly important for practical security evaluation because they remain effective even when the target model is inaccessible. In this setting, compared with conventional $L_p$-bounded attacks~\cite{goodfellow2014explaining,madry2017towards,dong2018boosting}, unrestricted adversarial transfer attacks~\cite{dai2024advdiff,xu2024highly,pan2024sca,Chen2023ContentbasedUA,DiffAttack} offer a more promising direction by relaxing pixel-space constraints and enabling semantically plausible manipulations, thereby improving transferability and imperceptibility.

Recently, diffusion models have emerged as a powerful foundation for unrestricted attacks~\cite{DiffAttack,li2023diffusion,dai2025semdiff}. By optimizing perturbations along latent denoising trajectories, diffusion-based attacks can generate adversarial examples with stronger semantic coherence and better transferability. However, existing methods still suffer from two fundamental limitations.

\begin{figure}[t]
    \centering
\includegraphics[width=1\linewidth]{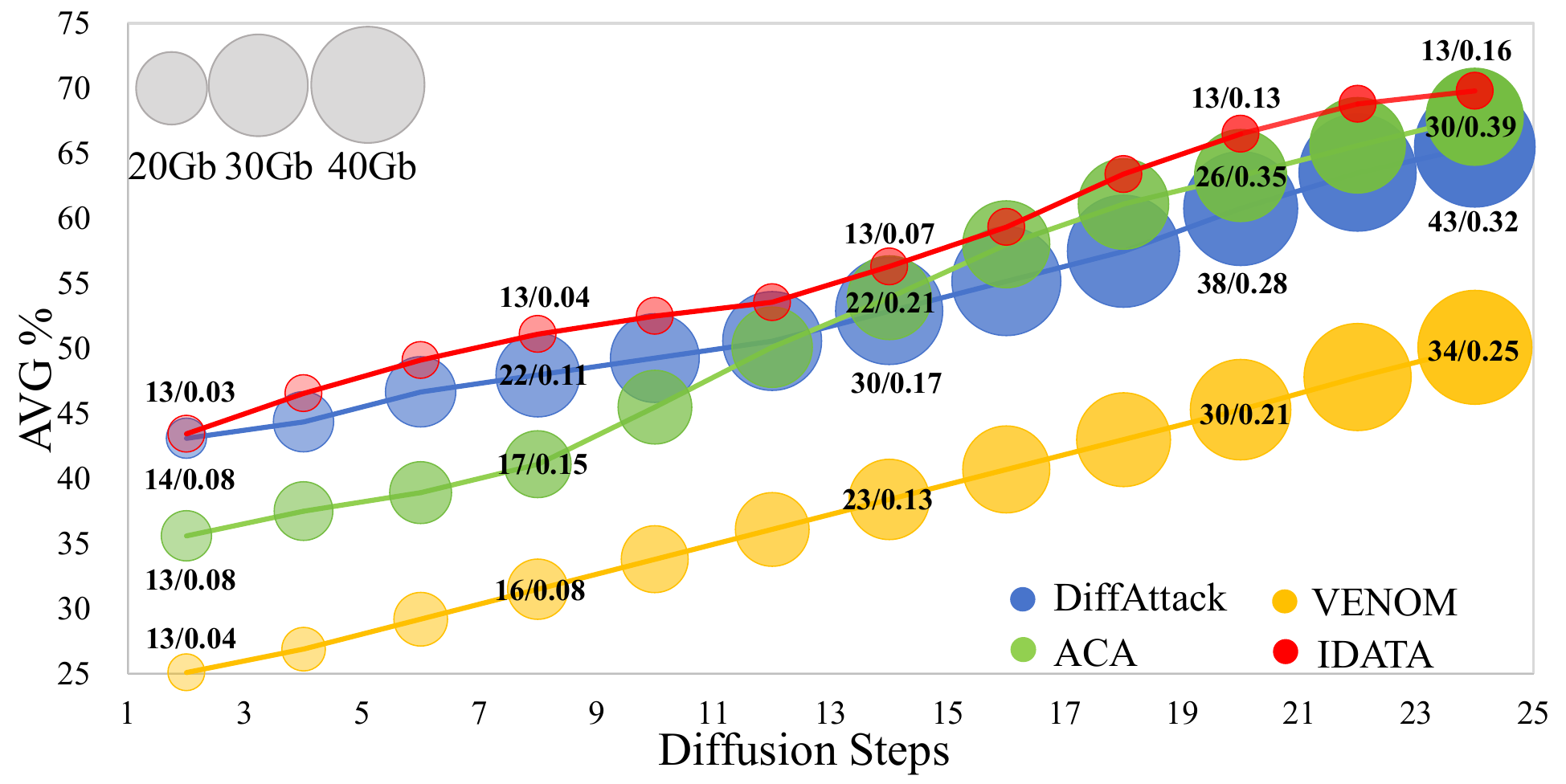}
    \caption{Comparison of DiffAttack~\cite{DiffAttack}, VENOM~\cite{kuurila2025venom}, ACA~\cite{Chen2023ContentbasedUA}, and IDATA at different diffusion depths. Circle size represents peak GPU memory usage (GB), while color intensity indicates LPIPS (darker means higher). The ``value/value'' label denotes GPU memory usage and LPIPS, respectively.}
    \label{pic:fig1}
\end{figure}

\textbf{First, trajectory-level optimization is prohibitively memory intensive.}
Current methods~\cite{xu2024highly, pan2024sca, Chen2023ContentbasedUA,kuurila2025venom,DiffAttack} require backpropagation through multiple denoising steps and must store intermediate states across the diffusion chain. As diffusion depth increases, memory cost grows rapidly, making deep optimization impractical. As shown in Fig.~\ref{pic:fig1}, although deeper trajectories generally improve attack performance, they also incur substantially higher GPU memory overhead. This poor scalability often forces prior methods to use shallow diffusion trajectories, thereby severely limiting the depth that can be practically optimized.

\textbf{Second, latent perturbation is largely frequency-unaware.}
Most existing diffusion-based attacks~\cite{li2023diffusion,dai2025semdiff,dai2024advdiff,xu2024highly,pan2024sca,Chen2023ContentbasedUA,DiffAttack} optimize perturbations over the full latent space without considering frequency structure, although intermediate diffusion latents exhibit timestep-dependent frequency characteristics. As shown in Fig.~\ref{pic:visualization}, such full-space perturbation tends to introduce redundant high-frequency artifacts and may overlook semantically stable structures shared across models. Prior studies~\cite{fan2025exploiting,10.5555/3367471.3367512,zhao2024improving,zhu2024frequency,zhu2025guided} suggest that low-frequency perturbations are generally more transferable, as they better preserve stable semantic content and are less sensitive to architecture-specific biases. However, incorporating such frequency-aware priors into intermediate diffusion latents remains largely unexplored.

These observations suggest that effective diffusion-based transfer attacks should jointly address \emph{deep trajectory optimization} and \emph{frequency-aware perturbation allocation}. To this end, we propose \textbf{IDATA}, short for \textbf{I}nvertible \textbf{D}iffusion for Unrestricted \textbf{A}dversarial \textbf{T}ransfer \textbf{A}ttack. IDATA consists of two tightly coupled components. First, an Invertible Diffusion Module (IDM) reformulates adversarial optimization as an invertible diffusion process, enabling exact reverse reconstruction of intermediate states and thus constant-memory backpropagation regardless of diffusion depth. Second, a Low-Frequency Constraint Module (LFCM) decomposes intermediate latents and constrains perturbations to semantically stable low-frequency subspaces. Together, these two modules enable scalable deep optimization and more transferable perturbation learning while preserving visual imperceptibility.

The main contributions of this work are summarized as follows:
\begin{itemize}
    \item[$\bullet$] \noindent We identify two critical bottlenecks in diffusion-based adversarial transfer attacks: memory-intensive multistep backpropagation and frequency-agnostic perturbation over intermediate latents, which collectively limit their scalability and transferability.
    \item[$\bullet$] \noindent To address these issues, we propose \textbf{IDATA}, a novel and scalable framework. It features an \textbf{Invertible Diffusion Module (IDM)} that enables constant-memory backpropagation, and a \textbf{Low-Frequency Constraint Module (LFCM)} that explicitly regularizes perturbations in semantically stable low-frequency latent subspaces.
    \item[$\bullet$] \noindent Extensive experiments on multiple architectures, defenses, and datasets show that IDATA consistently outperforms state-of-the-art baselines, achieving a favorable balance between transferability and visual imperceptibility.
\end{itemize}

\begin{figure*}[t]
    \centering
\includegraphics[width=0.985\linewidth]{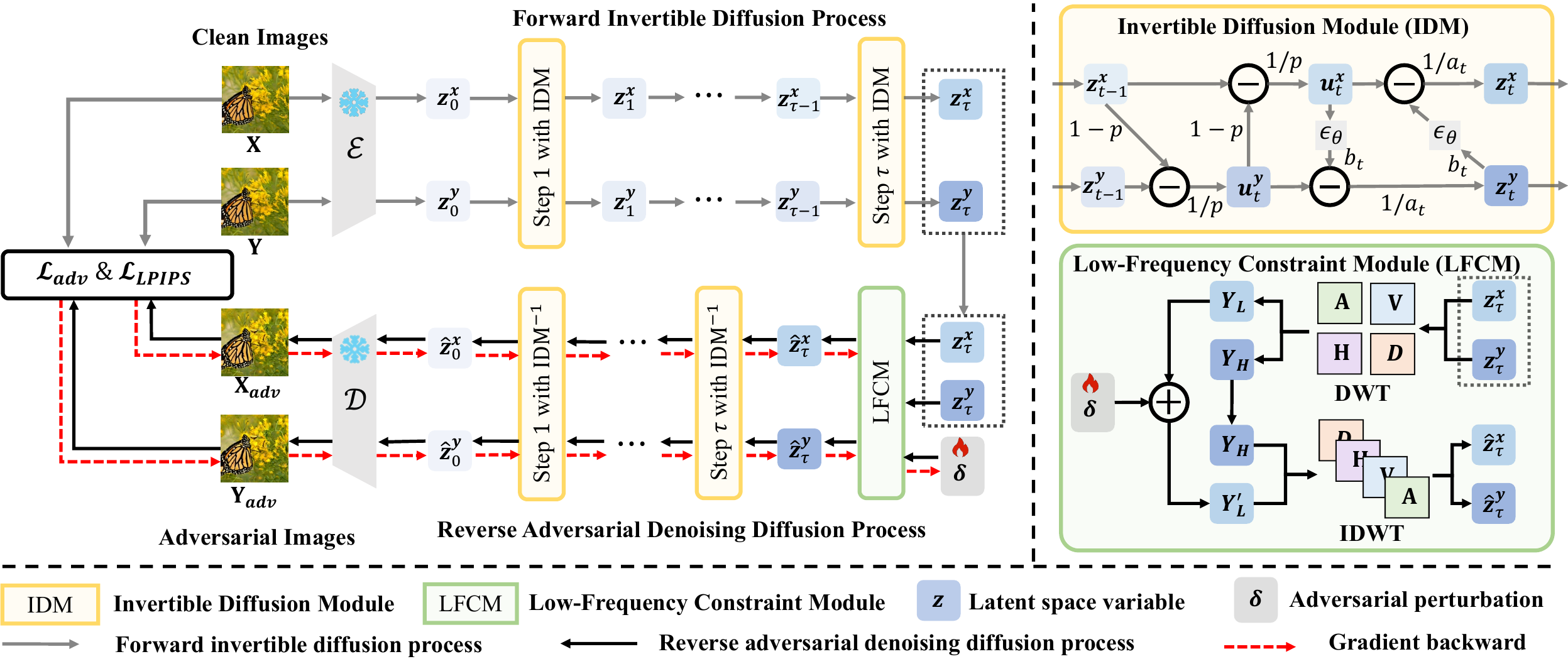}
    \caption{{Overview of} Invertible Diffusion Adversarial Transfer Attack (IDATA) framework. It consists of two core modules: the Invertible Diffusion Module (IDM) and the Low-Frequency Constraint Module (LFCM). Specifically, a clean image $\mathbf{X}$ is duplicated and processed by the forward IDM up to step $\tau$. The LFCM then injects a learnable perturbation $\bm{\delta}$ into the low-frequency components of the latent pair. Subsequently, the latents are reconstructed via the inverse process, $\text{IDM}^{-1}$. Crucially, IDM's perfect invertibility enables constant-memory gradient backpropagation for efficient optimization of $\bm{\delta}$. Finally, the primary branch is decoded to produce the adversarial image $\mathbf{X}_{\text{adv}}$.}
    \label{pic:overview}
\end{figure*}

\section{Related Work}
We review prior work from three perspectives relevant to our method and highlight the research gaps addressed in this paper.

\subsection{Unrestricted Adversarial Transfer Attack}
Adversarial transfer attacks are widely used for black-box robustness evaluation, as they require only a surrogate model to attack unknown targets~\cite{dong2018boosting,xie2019improving}. Early methods mainly focused on $L_p$-bounded pixel-space perturbations, such as MI-FGSM~\cite{dong2018boosting}, DI-FGSM~\cite{xie2019improving}, and PI-FGSM~\cite{gao2020patch}. Although effective in white-box settings, they often show limited transferability in black-box settings due to overfitting to surrogate decision boundaries and sensitivity to common preprocessing operations.

To boost transferability, several studies introduced frequency-domain priors. SSA~\cite{long2022frequency} enhances attack diversity via DCT/IDCT-based spectrum simulation, FRM~\cite{li2023frequency} improves generalization via low-frequency regularization, and SSAH~\cite{luo2022frequency} further considers semantic similarity, imperceptibility, and cross-model transferability. However, these still operate under rigid pixel-space $L_p$ constraints, limiting their ability to exploit semantic-level perturbation spaces.

To address this, unrestricted adversarial transfer attacks relax strict pixel-wise constraints, yielding semantically plausible examples with stronger transferability and better imperceptibility. Early methods like ReColorAdv~\cite{NEURIPS2019_6e923226} and cAdv~\cite{bhattad2019unrestricted} rely on handcrafted transformations with limited expressiveness, motivating more powerful generative approaches.

\subsection{Diffusion-based Adversarial Transfer Attack}
Diffusion models have become a promising foundation for unrestricted adversarial transfer attacks, as they can generate semantically coherent and visually plausible adversarial examples by perturbing latent denoising trajectories. ACA~\cite{Chen2023ContentbasedUA} first explored this paradigm by optimizing perturbations on the latent manifold of pre-trained diffusion models. DiffAttack~\cite{DiffAttack} improved visual imperceptibility through cross-attention perturbation with self-attention consistency, while VENOM~\cite{kuurila2025venom}, LESEP~\cite{gan2024lesep}, and DiffAdvMAP~\cite{pan2025diffadvmap} further enhanced attack flexibility via text-guided generation, semantic perturbation, and latent-space mapping.
Despite their effectiveness, these methods generally require backpropagation through multiple denoising steps, which incurs substantial memory overhead and limits their scalability for deep trajectory optimization.

While frequency priors have shown promise in conventional attacks~\cite{fan2025exploiting,10.5555/3367471.3367512,zhao2024improving,zhu2024frequency}, their use in diffusion-based attacks remains relatively underexplored. For example, LGA~\cite{zhu2025guided} constrains perturbations to low-frequency image components. Nevertheless, such methods primarily operate in the pixel space, leaving the distinct frequency characteristics of intermediate diffusion latents largely unexplored. These limitations, namely poor scalability and the lack of frequency-aware latent perturbation, motivate our work.

\subsection{Invertible Neural Networks}
Invertible Neural Networks (INNs)~\cite{huang2022winnet, huang2021linn} offer lossless bidirectional computation and constant-memory backpropagation via exact invertible transformations. Early affine coupling architectures, \textit{e.g.}, RealNVP~\cite{dinh2016density} and Glow~\cite{kingma2018glow}, demonstrated that invertibility preserves representational power while maintaining memory efficiency, which has also benefited adversarial attacks~\cite{chen2023imperceptible, pan2024svastin, chen2024invertible, huang2025smilenet}.

More recently, invertible diffusion pipelines have been integrated to boost reconstruction accuracy and optimization efficiency. EDICT~\cite{wallace2023edict} achieves exact reconstruction via coupled affine transformations, circumventing the error accumulation inherent to standard DDIM inversion. Subsequent extensions include an invertible diffusion framework~\cite{chen2025invertible} for compressed sensing with reduced memory and steps, and DOODL~\cite{wallace2023end} for end-to-end latent optimization that improves classifier-guided generation. Together, these results indicate the potential of invertible diffusion for efficient optimization in diffusion-based models.

However, existing invertible diffusion methods are not designed for unrestricted adversarial transfer attacks and have not been adapted to address attack-specific optimization challenges, particularly memory-efficient multistep optimization along diffusion trajectories. To address this gap, we propose IDATA.

\section{Preliminary}
\label{sebsec:DTAA}
We first revisit diffusion-based adversarial transfer attacks.
Such attacks aim to generate an adversarial example $\mathbf{X}_{adv}$ that misleads an unseen target classifier $\bm{f}_{\phi}(\cdot)$, which is structurally different from the surrogate classifier $\bm{f}_{\theta}(\cdot)$. Diffusion-based adversarial attacks have recently emerged as a promising paradigm for this task, as diffusion models provide a powerful generative prior for producing semantically coherent and visually realistic adversarial examples.

\textbf{Forward Diffusion Process:}
A clean sample $\mathbf{X}$ is first encoded to the initial latent $\bm{z}_0 = \mathcal{E}(\mathbf{X})$, which then undergoes $T$-step forward diffusion. For deterministic traversal, the DDIM framework is commonly used, with sampling:
$\bm{z}_t = \sqrt{\bar{\alpha}_t} \bm{z}_0 + \sqrt{1 - \bar{\alpha}_t} \epsilon, \text{where } \epsilon \sim \mathcal{N}(0, \mathbf{I})$, and $\bar{\alpha}_{t} = \prod_{i=1}^{t} \alpha_{i}$ is the cumulative product of noise schedule coefficients up to step $t$, corresponding to the overall signal retention rate.

\textbf{Reverse Denoising Diffusion Process:}
In denoising, the early steps establish high-level semantics while later steps recover fine details.
Prior diffusion-based attacks often perturb latent variables at early steps $(\tau \ll T)$, \textit{e.g.}, $\tau = \frac{1}{10}T$, to avoid gradient explosion and preserve optimization stability.
Specifically, the adversarial perturbation is introduced as $\hat{\bm{z}}_{\tau}^{adv} = \bm{z}_{\tau} + \bm{\delta}$, where $\bm{\delta}$ denotes the learnable adversarial latent perturbation.
The adversarial latent is then refined via iterative denoising: $\hat{\bm{z}}_{t-1}^{adv} = a_t{\hat{\bm{z}}}_{t}^{adv} + b_t {\epsilon}_\theta(\hat{\bm{z}}_t^{{adv}}, t, C),$
where $a_t = \sqrt{\bar{\alpha}_{t-1}/\bar{\alpha}_t}$, $b_t = \sqrt{1-\bar{\alpha}_{t-1}} - \sqrt{\bar{\alpha}_{t-1}(1-\bar{\alpha}_t)/\bar{\alpha}_t}$, $C$ denotes conditional information (\textit{e.g.}, text annotations), and ${\epsilon}_\theta(\cdot, t, C)$ estimates step-$t$ noise under $C$.

The adversarial image is finally obtained as $\mathbf{X}_{adv} = \mathcal{D}(\hat{\bm{z}}_0^{adv})$, where $\mathcal{D}(\cdot)$ denotes the decoder of the Variational Auto-Encoder (VAE), which takes the reverse-updated adversarial latent code $\hat{\bm{z}}_0^{adv}$ as input. The adversarial latent $\bm{\delta}$ is optimized through gradient backpropagation in the reverse denoising process by minimizing the adversarial loss $\mathcal{L}(\cdot)$ with respect to $\bm{\delta}$. 

However, backpropagation through the reverse denoising trajectory incurs substantial memory overhead, which directly limits the number of feasible diffusion steps during optimization. As illustrated in Fig.~\ref{pic:fig1}, existing diffusion-based adversarial attacks exhibit memory complexity that increases approximately linearly with the number of perturbation steps $\tau$. This memory bottleneck severely constrains the scalability of diffusion-based attacks, especially for high-resolution multimedia inputs.

\section{Methodology}
We present \emph{\textbf{I}nvertible \textbf{D}iffusion for unrestricted \textbf{A}dversarial \textbf{T}ransfer \textbf{A}ttack} (IDATA). As shown in Fig.~\ref{pic:overview}, IDATA includes two components: 1) an Invertible Diffusion Module (IDM) for memory-efficient adversarial optimization over deep diffusion trajectories; and 2) a Low-Frequency Constraint Module (LFCM) that restricts perturbations to semantically stable low-frequency latent subspaces, improving transferability while preserving visual imperceptibility.

\subsection{Invertible Diffusion Module}
To overcome the memory bottleneck in diffusion-based attacks, we introduce the \emph{Invertible Diffusion Module} (IDM), which enables adversarial optimization with a constant memory footprint, regardless of the diffusion trajectory depth.
Notably, while IDM is inspired by the invertible design of EDICT~\cite{wallace2023edict}, 
it is fundamentally redesigned for adversarial optimization rather than faithful 
reconstruction.

\textbf{Forward Invertible Diffusion:}
The IDM process is formulated using a primary and an auxiliary branch. A single clean image $\mathbf{X}$ is duplicated to form an input pair $(\mathbf{X}, \mathbf{Y=X})$ to satisfy the coupled-branch requirement of the invertible transformations. This pair is then encoded via the VAE's encoder into initial latents $\bm{z}_0^{\bm{x}} = \mathcal{E}(\mathbf{X})$ and $\bm{z}_0^{\bm{y}} = \mathcal{E}(\mathbf{Y})$, where the $\bm{z}^{\bm{y}}$ branch serves as the auxiliary component necessary for invertibility.
Each invertible diffusion step employs two coupled transformations. In forward diffusion at step $t$, the latent pair $(\bm{z}_{t-1}^{\bm{x}}, \bm{z}_{t-1}^{\bm{y}})$ is mixed using a stability parameter $p \in (0,1]$ to produce the intermediate latent pair $\left(\bm{u}_{t}^{\bm{x}},\bm{u}_{t}^{\bm{y}}\right)$:
\begin{align}
\small
    \begin{cases}
        \bm{u}_{t}^{\bm{y}} &= \left( \bm{z}_{t-1}^{\bm{y}} - (1-p) \bm{z}_{t-1}^{\bm{x}} \right)/p, \\
        \bm{u}_{t}^{\bm{x}} &= \left( \bm{z}_{t-1}^{\bm{x}} - (1-p) \bm{u}_{t}^{\bm{y}} \right)/p.
    \end{cases} 
\label{eq:eq1}
\end{align}
\noindent Then, forward scheduled noise addition is applied to these mixed features using coefficients $a_{t}, b_{t} \in \mathbb{R}$ and the noise estimator $\epsilon_{\theta}$:
\begin{align}
\small
\begin{cases}
    \bm{z}_{t}^{\bm{y}} &= \left(\bm{u}_{t}^{\bm{y}} - b_{t} \, \epsilon_{\theta}(\bm{u}_{t}^{\bm{x}}, t, C)\right) / a_{t}, \\
    \bm{z}_{t}^{\bm{x}} &= \left(\bm{u}_{t}^{\bm{x}} - b_{t} \, \epsilon_{\theta}(\bm{z}_{t}^{\bm{y}}, t, C)\right) / a_{t}.
\end{cases}
\label{eq:eq2}
\end{align}
This sequential parameterization is critical for invertibility, as each variable can be recovered in closed form during the reverse process.

\textbf{Reverse Adversarial Denoising Diffusion:}
Given the forward IDM in Eqn.~\eqref{eq:eq1} and ~\eqref{eq:eq2}, the reverse process reconstructs the previous latent pair $(\hat{\bm{z}}_{t-1}^{\bm{x}}, \hat{\bm{z}}_{t-1}^{\bm{y}})$ from the current noisy pair $(\hat{\bm{z}}_{t}^{\bm{x}}, \hat{\bm{z}}_{t}^{\bm{y}})$ by exactly inverting the two coupled transformations at each timestep.
At the perturbation injection step $\tau \in [1,T]$, we introduce a learnable perturbation $\bm{\delta}$ which is integrated into the low-frequency components of the latent pair via a Low-Frequency Constraint Module (LFCM) to balance transferability and imperceptibility. The LFCM will be detailed in Sec.~\ref{sec:LFCM}. Formally, $(\hat{\bm{z}}_{\tau}^{\bm{x}}, \hat{\bm{z}}_{\tau}^{\bm{y}}) = \text{LFCM}(\bm{z}_{\tau}^{\bm{x}},\bm{z}_{\tau}^{\bm{y}}, \bm{\delta})$.

Since gradients propagated through deep diffusion trajectories may exhibit high variance, directly optimizing $\bm{\delta}$ can lead to unstable updates. To alleviate this issue, we adopt the Momentum Iterative Fast Gradient Sign Method (MI-FGSM)~\cite{dong2018boosting} to stabilize adversarial latent optimization. Given the latent pair $(\bm{\hat{z}}_{\tau}^{x}, \bm{\hat{z}}_{\tau}^{y})$ at denoising step $\tau$, the momentum term at the $i$-th attack iteration is computed as
\begin{equation}
\small
\bm{m}_{i+1} = \mu \cdot \bm{m}_i + 
\frac{\nabla_{\bm{\delta}_i} \mathcal{L}}
{\|\nabla_{\bm{\delta}_i} \mathcal{L}\|_1},
\label{eq:eq11}
\end{equation}
where $\mathcal{L}$ denotes the loss function and the perturbation $\bm{\delta}$ is updated by
\begin{equation}
\small
\bm{\delta}_{i+1} = \bm{\delta}_i + \eta \cdot \mathrm{sign}(\bm{m}_{i+1}),
\label{eq:eq9}
\end{equation}
where $\mu$ and $\eta$ denote the momentum decay factor and step size, respectively. This update smooths timestep-wise gradient fluctuations and improves optimization stability. Following the deterministic reverse formulation, each reverse step then proceeds in two stages:

1) Inverse Scheduled Denoising:
We recover the intermediate variables $(\bm{u}_{t}^{\bm{x}}, \bm{u}_{t}^{\bm{y}})$ by inverting the affine update in Eqn.~\eqref{eq:eq2}:
\begin{align}
\small
\begin{cases}
    \bm{u}_{t}^{\bm{x}} = a_t \hat{\bm{z}}_{t}^{\bm{x}} + b_t \, \epsilon_{\theta}(\hat{\bm{z}}_{t}^{\bm{y}}, t, C), \\
    \bm{u}_{t}^{\bm{y}} = a_t \hat{\bm{z}}_{t}^{\bm{y}} + b_t \, \epsilon_{\theta}(\bm{u}_{t}^{\bm{x}}, t, C).
\end{cases}
\label{eq:eq3}
\end{align}

2) Inverse Feature Mixing:
We then invert Eqn.~\eqref{eq:eq1} to obtain the previous latent pair $(\hat{\bm{z}}_{t-1}^{\bm{x}}, \hat{\bm{z}}_{t-1}^{\bm{y}})$:
\begin{align}
\small
\begin{cases}
    \hat{\bm{z}}_{t-1}^{\bm{x}} = p \bm{u}_{t}^{\bm{x}} + (1-p)\bm{u}_{t}^{\bm{y}}, \\
    \hat{\bm{z}}_{t-1}^{\bm{y}} = p \bm{u}_{t}^{\bm{y}} + (1-p)\hat{\bm{z}}_{t-1}^{\bm{x}}.
\end{cases}
\label{eq:eq4}
\end{align}
After the reverse process reaches $t=0$, the outputs of the two branches are decoded as
$\mathbf{X}_{adv}=\mathcal{D}(\hat{\bm{z}}_{0}^{\bm{x}})$ and
$\mathbf{Y}_{adv}=\mathcal{D}(\hat{\bm{z}}_{0}^{\bm{y}})$,
where $\mathcal{D}(\cdot)$ denotes the VAE decoder. We use $\mathbf{X}_{adv}$ as the final adversarial example.

As shown in Fig.~\ref{pic:fig1}, unlike conventional attacks with memory usage scaling linearly with optimization steps, IDM's invertibility allows it to dynamically reconstruct intermediate states during backpropagation. This reduces the activation memory required for trajectory backpropagation to $\mathcal{O}(1)$ with respect to diffusion depth and unlocks efficient optimization over deep diffusion trajectories.

\paragraph{Difference from EDICT}
IDM is related to EDICT~\cite{wallace2023edict} but differs substantially in goal and functionality. EDICT focuses on exact diffusion inversion and faithful reconstruction through coupled invertible transformations, mainly to reduce DDIM inversion errors. By contrast, IDM is designed for unrestricted adversarial transfer attacks, where invertibility is exploited to support exact constant-memory backpropagation and attack-oriented latent optimization. In addition, IDM is coupled with perturbation injection during adversarial optimization, which is beyond the scope of EDICT.

\subsection{Low-Frequency Constraint Module}
\label{sec:LFCM}
While IDM enables memory-efficient optimization over deep diffusion trajectories, strong attack transferability also depends critically on how perturbations are allocated in the latent space. Existing diffusion-based attack methods~\cite{kuurila2025venom,DiffAttack,pan2025diffadvmap} largely overlook this issue and perturb the latent space indiscriminately, which often introduces visible high-frequency artifacts and undermines semantic transferability, as illustrated in Fig.~\ref{pic:visualization} and Fig.~\ref{pic:ablation_vis}. To address this limitation, we propose a \emph{Low-Frequency Constraint Module} (LFCM) that restricts perturbations to the semantically stable low-frequency subspace at intermediate timesteps. This design improves cross-model transferability while better preserving visual imperceptibility.

\begin{figure}[t]
    \centering
    \includegraphics[width=1\linewidth]{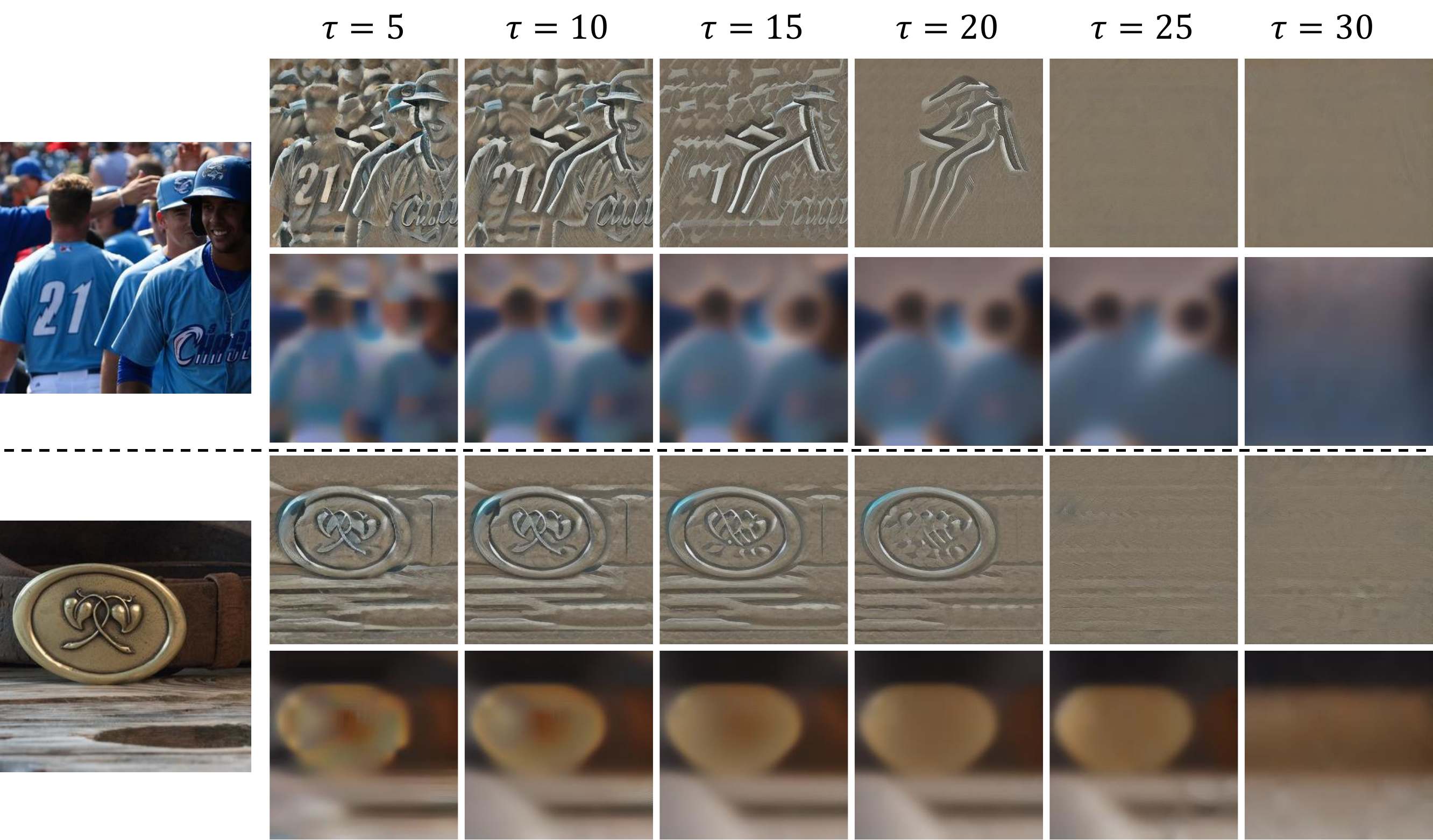}
    \caption{Reconstructions from high- and low-frequency components of diffusion latents at timesteps $\tau$. The leftmost column shows clean images. The top row uses only high-frequency components, and the bottom row uses only low-frequency components, with the other component set to $0$. Results are shown with $T=50$.}
    \label{pic:high_low}
\end{figure}

To motivate this design, we analyze the frequency characteristics of intermediate latent variables across timesteps, as shown in Fig.~\ref{pic:high_low}. Low-frequency components preserve global structure, object shape, and core semantics throughout denoising, while high-frequency components capture local textures whose semantic validity decays rapidly as timestep $\tau$ increases. Full-latent perturbation thus introduces redundant high-frequency noise and wastes budget on semantically fragile components, degrading both imperceptibility and cross-model transferability.

Based on this observation, at diffusion step $\tau$, we first concatenate the intermediate latent variables from both branches along the channel dimension to form a unified representation, $\bm{z}_{\tau} = [\bm{z}_{\tau}^{x}, \bm{z}_{\tau}^{y}]$. The LFCM then decomposes this unified latent $\bm{z}_{\tau}$ into its low-frequency and high-frequency components.
To this end, we employ the Discrete Wavelet Transform (DWT)~\cite{mallat2002theory} with Haar wavelet basis to implement this decomposition. Specifically, given a latent representation $\bm{z}_{\tau} \in \mathbb{R}^{B \times C \times H \times W}$, where $B$ denotes the batch size and $C$ represents the number of channels, we use a low-pass filter $\bm{L}$ and a high-pass filter $\bm{H}$ to disentangle $\bm{z}_{\tau}$ into a low-frequency component $\boldsymbol{Y}_{L} \in \mathbb{R}^{B \times C \times \frac{H}{2} \times \frac{W}{2}}$ and a high-frequency component $\boldsymbol{Y}_{H} \in \mathbb{R}^{B \times 3C \times \frac{H}{2} \times \frac{W}{2}}$:
\begin{equation}
\small
    (\boldsymbol{Y}_{L}, \boldsymbol{Y}_{H}) = \mathrm{DWT}(\bm{z}_{\tau}),
    \label{eq:eq7}
\end{equation}
where $\boldsymbol{Y}_{L} = \bm{L} \bm{z}_{\tau} \bm{L}^{T}$ denotes the low-frequency approximation, and $\boldsymbol{Y}_{H} = [\boldsymbol{c}_{hl}, \boldsymbol{c}_{lh}, \boldsymbol{c}_{hh}] 
= [\bm{H} \bm{z}_{\tau} \bm{L}^{T}, \bm{L} \bm{z}_{\tau} \bm{H}^{T}, \bm{H} \bm{z}_{\tau} \bm{H}^{T}]$ concatenates the detailed high-frequency information across different spatial orientations.
Subsequently, the Inverse Discrete Wavelet Transform (IDWT) is applied to $\boldsymbol{Y}_{L}$ and $\boldsymbol{Y}_{H}$ to recover the original latent variable $\bm{z}_{\tau}$: $\bm{z}_{\tau} = \text{IDWT}(\boldsymbol{Y}_{L}, \boldsymbol{Y}_{H})$.

To improve transferability without sacrificing visual quality, we inject the adversarial perturbation only into the low-frequency component, \textit{i.e.}, $\boldsymbol{Y}_{L}^{\prime} = \boldsymbol{Y}_{L} + \bm{\delta}$, while keeping $\boldsymbol{Y}_{H}$ unchanged. The perturbed latent is reconstructed as $\bm{z}_{\tau}^{\prime} = \text{IDWT}(\boldsymbol{Y}^{\prime}_{L}, \boldsymbol{Y}_{H})$.
Crucially, to interface back with the dual-branch IDM, we then split $\bm{z}_{\tau}^{\prime}$ along the channel dimension to obtain the perturbed latent pair $(\hat{\bm{z}}_{\tau}^{x}, \hat{\bm{z}}_{\tau}^{y})$. This pair is then fed into the subsequent reverse diffusion process.
In this way, LFCM serves as a frequency-aware latent regularizer that steers optimization toward model-agnostic global semantics rather than surrogate-specific high-frequency details. Compared with full-latent perturbation, LFCM suppresses high-frequency artifacts, improves perturbation budget utilization, and achieves a better transferability-imperceptibility trade-off across heterogeneous architectures and defenses. Moreover, LFCM is lightweight, model-agnostic, and plug-and-play, readily compatible with existing diffusion-based attacks without additional fine-tuning.

\subsection{Loss Function}
Integrating the IDM and LFCM modules, our final objective is designed to jointly enforce cross-model transferability and visual imperceptibility. The overall loss function combines an adversarial guidance loss $\mathcal{L}_{\text{adv}}$ with a perceptual quality loss $\mathcal{L}_{\text{LPIPS}}$:
\begin{align}
    \mathcal{L}_{\text{total}} =  \mathcal{L}_{{adv}} +  \lambda
    \mathcal{L}_{\text{LPIPS}},
\label{eq:total_loss}
\end{align}
\noindent where $\lambda$ is a regularization parameter.
The adversarial guidance loss $\mathcal{L}_{\text{adv}}$ is computed symmetrically for both the primary ($\mathbf{X}_{\text{adv}}$) and auxiliary ($\mathbf{Y}_{\text{adv}}$) branches to enforce trajectory synchronization within the IDM. This co-optimization strategy explicitly couples their adversarial paths and is defined as: $\mathcal{L}_{{adv}} = -\mathcal{L}_{CE}\left(\mathbf{X}_{{adv}}, y; \bm{f}_{\theta}\right) -\mathcal{L}_{CE}\left(\mathbf{Y}_{{adv}}, y; \bm{f}_{\theta}\right)$, where $\mathcal{L}_{CE}(\cdot)$ denotes the cross-entropy loss and $y$ is the image's true label.
The perceptual quality loss $\mathcal{L}_{\text{LPIPS}}$ is defined as $\mathcal{L}_{\text{LPIPS}} = \text{LPIPS}(\mathbf{X}, \mathbf{X}_{{adv}}) + \text{LPIPS}(\mathbf{Y}, \mathbf{Y}_{{adv}})$, which preserves perceptual similarity between clean and adversarial images.

\begin{table*}[t]
  \renewcommand{\arraystretch}{0.5}
  \centering
  \caption{Comparison of transferability and imperceptibility on normally trained models. We report top-1 accuracy (\%) of each method. ``S'' denotes surrogate models, and ``T'' denotes target models. ``AVG w/o self'' represents the average ASR (Attack Success Rate) over all target models excluding the surrogate model itself. Different shades of blue indicate ASR values on target models with different architectures, where darker blue denotes higher ASR. The best results are highlighted in bold \textcolor{red}{red} font.}
  \resizebox{0.984\textwidth}{!}{
    \begin{tabular}{cc|ccccc|cccc|cc|c|cc}
    \toprule
    \multirow{3}[6]{*}{\diagbox[width=2em]{S.}{T.}} & \multirow{2}[4]{*}{Attacks} & \multicolumn{5}{c|}{CNNs}             & \multicolumn{4}{c|}{Transformers} & \multicolumn{2}{c|}{MLPs} & \multicolumn{1}{c|}{\multirow{2}[4]{*}{\thead{AVG \\ (w/o self)}$\uparrow$}} & \multirow{2}[4]{*}{FID$\downarrow$} & \multirow{2}[4]{*}{LPIPS$\downarrow$} \\
\cmidrule{3-13}          &       & Res-50 & VGG-19 & Mob-v2 & Inc-v3 & ConvNeXt & Vit-B & Swin-B & DeiT-B & Deit-S & Mix-B & Mix-L &       &       &  \\
\cmidrule{2-16}          & Clean & 7.3   & 11.3  & 13.1  & 19.5  & 3     & 6.3   & 4.1   & 5.5   & 6     & 17.5  & 23.5  & 10.6  & 57.8  & - \\
    \midrule
    \midrule
    \multirow{9}[2]{*}{Res-50} & DI-FGSM & \cellcolor[rgb]{ .965,  .969,  .988}100  & \cellcolor[rgb]{ .965,  .969,  .988}75.8  & \cellcolor[rgb]{ .965,  .969,  .988}75.5  & \cellcolor[rgb]{ .965,  .969,  .988}65.5  & \cellcolor[rgb]{ .965,  .969,  .988}26.4  & \cellcolor[rgb]{ .929,  .949,  .976}18.4  & \cellcolor[rgb]{ .929,  .949,  .976}17.7  & \cellcolor[rgb]{ .929,  .949,  .976}19.5  & \cellcolor[rgb]{ .929,  .949,  .976}24.0  & \cellcolor[rgb]{ .886,  .91,  .965}38.7  & \cellcolor[rgb]{ .886,  .91,  .965}43.2  & 40.5  & 85.3  & 0.143  \\
          & PI-FGSM & \cellcolor[rgb]{ .965,  .969,  .988}100  & \cellcolor[rgb]{ .965,  .969,  .988}75.9  & \cellcolor[rgb]{ .965,  .969,  .988}75.0  & \cellcolor[rgb]{ .965,  .969,  .988}76.0  & \cellcolor[rgb]{ .965,  .969,  .988}27.5  & \cellcolor[rgb]{ .929,  .949,  .976}34.7  & \cellcolor[rgb]{ .929,  .949,  .976}22.5  & \cellcolor[rgb]{ .929,  .949,  .976}34.3  & \cellcolor[rgb]{ .929,  .949,  .976}35.0  & \cellcolor[rgb]{ .886,  .91,  .965}49.7  & \cellcolor[rgb]{ .886,  .91,  .965}56.2  & 48.7  & 97.9  & 0.136  \\
          & ReColorAdv & \cellcolor[rgb]{ .965,  .969,  .988}99.2  & \cellcolor[rgb]{ .965,  .969,  .988}57.6  & \cellcolor[rgb]{ .965,  .969,  .988}63.8  & \cellcolor[rgb]{ .965,  .969,  .988}47.9  & \cellcolor[rgb]{ .965,  .969,  .988}28.6  & \cellcolor[rgb]{ .929,  .949,  .976}31.7  & \cellcolor[rgb]{ .929,  .949,  .976}28.1  & \cellcolor[rgb]{ .929,  .949,  .976}26.8  & \cellcolor[rgb]{ .929,  .949,  .976}32.3  & \cellcolor[rgb]{ .886,  .91,  .965}36.5  & \cellcolor[rgb]{ .886,  .91,  .965}40.8  & 39.4  & 63.4  & 0.154  \\
          & cAdv  & \cellcolor[rgb]{ .965,  .969,  .988}97.6  & \cellcolor[rgb]{ .965,  .969,  .988}48.2  & \cellcolor[rgb]{ .965,  .969,  .988}57.5  & \cellcolor[rgb]{ .965,  .969,  .988}46.8  & \cellcolor[rgb]{ .965,  .969,  .988}21.0  & \cellcolor[rgb]{ .929,  .949,  .976}36.9  & \cellcolor[rgb]{ .929,  .949,  .976}24.7  & \cellcolor[rgb]{ .929,  .949,  .976}31.7  & \cellcolor[rgb]{ .929,  .949,  .976}37.7  & \cellcolor[rgb]{ .886,  .91,  .965}48.3  & \cellcolor[rgb]{ .886,  .91,  .965}56.0  & 40.9  & 65.7  & 0.156  \\
          & VENOM & \cellcolor[rgb]{ .965,  .969,  .988}100  & \cellcolor[rgb]{ .965,  .969,  .988}42.6  & \cellcolor[rgb]{ .965,  .969,  .988}43.3  & \cellcolor[rgb]{ .965,  .969,  .988}35.9  & \cellcolor[rgb]{ .965,  .969,  .988}68.8  & \cellcolor[rgb]{ .929,  .949,  .976}68.5  & \cellcolor[rgb]{ .929,  .949,  .976}62.3  & \cellcolor[rgb]{ .929,  .949,  .976}67.8  & \cellcolor[rgb]{ .929,  .949,  .976}66.0  & \cellcolor[rgb]{ .886,  .91,  .965}55.1  & \cellcolor[rgb]{ .886,  .91,  .965}45.8  & 55.6  & 80.2  & 0.169  \\
          & ACA   & \cellcolor[rgb]{ .965,  .969,  .988}88.0  & \cellcolor[rgb]{ .965,  .969,  .988}67.7  & \cellcolor[rgb]{ .965,  .969,  .988}71.5  & \cellcolor[rgb]{ .965,  .969,  .988}69.4  & \cellcolor[rgb]{ .965,  .969,  .988}53.0  & \cellcolor[rgb]{ .929,  .949,  .976}57.0  & \cellcolor[rgb]{ .929,  .949,  .976}54.9  & \cellcolor[rgb]{ .929,  .949,  .976}54.8  & \cellcolor[rgb]{ .929,  .949,  .976}55.1  & \cellcolor[rgb]{ .886,  .91,  .965}60.4  & \cellcolor[rgb]{ .886,  .91,  .965}60.8  & 60.5  & 76.6  & 0.378  \\
          & DiffAttack & \cellcolor[rgb]{ .965,  .969,  .988}96.8  & \cellcolor[rgb]{ .965,  .969,  .988}75.6  & \cellcolor[rgb]{ .965,  .969,  .988}77.1  & \cellcolor[rgb]{ .965,  .969,  .988}69.0  & \cellcolor[rgb]{ .965,  .969,  .988}59.0  & \cellcolor[rgb]{ .929,  .949,  .976}51.2  & \cellcolor[rgb]{ .929,  .949,  .976}56.2  & \cellcolor[rgb]{ .929,  .949,  .976}50.5  & \cellcolor[rgb]{ .929,  .949,  .976}55.0  & \cellcolor[rgb]{ .886,  .91,  .965}57.1  & \cellcolor[rgb]{ .886,  .91,  .965}57.8  & 60.9  & 62.6  & 0.137  \\
          & DiffAdvMAP & \cellcolor[rgb]{ .965,  .969,  .988}100  & \cellcolor[rgb]{ .965,  .969,  .988}77.7  & \cellcolor[rgb]{ .965,  .969,  .988}82.7  & \cellcolor[rgb]{ .965,  .969,  .988}67.1  & \cellcolor[rgb]{ .965,  .969,  .988}62.1  & \cellcolor[rgb]{ .929,  .949,  .976}48.6  & \cellcolor[rgb]{ .929,  .949,  .976}57.9  & \cellcolor[rgb]{ .929,  .949,  .976}49.1  & \cellcolor[rgb]{ .929,  .949,  .976}53.4  & \cellcolor[rgb]{ .886,  .91,  .965}54.4  & \cellcolor[rgb]{ .886,  .91,  .965}55.5  & 60.9  & 60.2  & 0.136  \\
          & IDATA(Our) & \cellcolor[rgb]{ .965,  .969,  .988}97.7  & \cellcolor[rgb]{ .965,  .969,  .988}78.8  & \cellcolor[rgb]{ .965,  .969,  .988}80.6  & \cellcolor[rgb]{ .965,  .969,  .988}71.3  & \cellcolor[rgb]{ .965,  .969,  .988}62.2  & \cellcolor[rgb]{ .929,  .949,  .976}53.7  & \cellcolor[rgb]{ .929,  .949,  .976}59.3  & \cellcolor[rgb]{ .929,  .949,  .976}50.9  & \cellcolor[rgb]{ .929,  .949,  .976}56.0  & \cellcolor[rgb]{ .886,  .91,  .965}55.9  & \cellcolor[rgb]{ .886,  .91,  .965}58.4  & \textcolor[rgb]{ 1,  0,  0}{\textbf{62.7 }} & \textcolor[rgb]{ 1,  0,  0}{\textbf{59.8 }} & \textcolor[rgb]{ 1,  0,  0}{\textbf{0.133 }} \\
    \midrule
    \multirow{9}[2]{*}{Inc-v3} & DI-FGSM & \cellcolor[rgb]{ .965,  .969,  .988}38.3  & \cellcolor[rgb]{ .965,  .969,  .988}42.6  & \cellcolor[rgb]{ .965,  .969,  .988}43.3  & \cellcolor[rgb]{ .965,  .969,  .988}99.8  & \cellcolor[rgb]{ .965,  .969,  .988}10.1  & \cellcolor[rgb]{ .929,  .949,  .976}15.4  & \cellcolor[rgb]{ .929,  .949,  .976}13.2  & \cellcolor[rgb]{ .929,  .949,  .976}13.3  & \cellcolor[rgb]{ .929,  .949,  .976}18.0  & \cellcolor[rgb]{ .886,  .91,  .965}31.6  & \cellcolor[rgb]{ .886,  .91,  .965}37.7  & 33.0  & 67.1  & 0.131  \\
          & PI-FGSM & \cellcolor[rgb]{ .965,  .969,  .988}41.7  & \cellcolor[rgb]{ .965,  .969,  .988}42.6  & \cellcolor[rgb]{ .965,  .969,  .988}51.7  & \cellcolor[rgb]{ .965,  .969,  .988}100  & \cellcolor[rgb]{ .965,  .969,  .988}19.0  & \cellcolor[rgb]{ .929,  .949,  .976}25.7  & \cellcolor[rgb]{ .929,  .949,  .976}19.7  & \cellcolor[rgb]{ .929,  .949,  .976}18.1  & \cellcolor[rgb]{ .929,  .949,  .976}22.0  & \cellcolor[rgb]{ .886,  .91,  .965}42.9  & \cellcolor[rgb]{ .886,  .91,  .965}42.5  & 38.7  & 92.5  & 0.137  \\
          & ReColorAdv & \cellcolor[rgb]{ .965,  .969,  .988}31.6  & \cellcolor[rgb]{ .965,  .969,  .988}35.8  & \cellcolor[rgb]{ .965,  .969,  .988}39.3  & \cellcolor[rgb]{ .965,  .969,  .988}98.4  & \cellcolor[rgb]{ .965,  .969,  .988}11.7  & \cellcolor[rgb]{ .929,  .949,  .976}19.1  & \cellcolor[rgb]{ .929,  .949,  .976}15.0  & \cellcolor[rgb]{ .929,  .949,  .976}14.7  & \cellcolor[rgb]{ .929,  .949,  .976}19.6  & \cellcolor[rgb]{ .886,  .91,  .965}26.6  & \cellcolor[rgb]{ .886,  .91,  .965}32.6  & 31.3  & 63.3  & 0.141  \\
          & cAdv  & \cellcolor[rgb]{ .965,  .969,  .988}23.1  & \cellcolor[rgb]{ .965,  .969,  .988}27.7  & \cellcolor[rgb]{ .965,  .969,  .988}29.7  & \cellcolor[rgb]{ .965,  .969,  .988}86.6  & \cellcolor[rgb]{ .965,  .969,  .988}8.9  & \cellcolor[rgb]{ .929,  .949,  .976}18.6  & \cellcolor[rgb]{ .929,  .949,  .976}14.3  & \cellcolor[rgb]{ .929,  .949,  .976}14.6  & \cellcolor[rgb]{ .929,  .949,  .976}15.9  & \cellcolor[rgb]{ .886,  .91,  .965}20.3  & \cellcolor[rgb]{ .886,  .91,  .965}35.2  & 26.8  & 62.2  & 0.135  \\
          & VENOM & \cellcolor[rgb]{ .965,  .969,  .988}35.7  & \cellcolor[rgb]{ .965,  .969,  .988}33.8  & \cellcolor[rgb]{ .965,  .969,  .988}32.3  & \cellcolor[rgb]{ .965,  .969,  .988}100  & \cellcolor[rgb]{ .965,  .969,  .988}41.0  & \cellcolor[rgb]{ .929,  .949,  .976}43.5  & \cellcolor[rgb]{ .929,  .949,  .976}44.0  & \cellcolor[rgb]{ .929,  .949,  .976}38.8  & \cellcolor[rgb]{ .929,  .949,  .976}36.2  & \cellcolor[rgb]{ .886,  .91,  .965}30.2  & \cellcolor[rgb]{ .886,  .91,  .965}25.3  & 41.9  & 73.8  & 0.214  \\
          & ACA   & \cellcolor[rgb]{ .965,  .969,  .988}39.6  & \cellcolor[rgb]{ .965,  .969,  .988}42.6  & \cellcolor[rgb]{ .965,  .969,  .988}52.0  & \cellcolor[rgb]{ .965,  .969,  .988}93.0  & \cellcolor[rgb]{ .965,  .969,  .988}38.0  & \cellcolor[rgb]{ .929,  .949,  .976}25.8  & \cellcolor[rgb]{ .929,  .949,  .976}27.5  & \cellcolor[rgb]{ .929,  .949,  .976}29.6  & \cellcolor[rgb]{ .929,  .949,  .976}31.3  & \cellcolor[rgb]{ .886,  .91,  .965}34.9  & \cellcolor[rgb]{ .886,  .91,  .965}45.6  & 41.8  & 80.4  & 0.390  \\
          & DiffAttack & \cellcolor[rgb]{ .965,  .969,  .988}40.4  & \cellcolor[rgb]{ .965,  .969,  .988}44.5  & \cellcolor[rgb]{ .965,  .969,  .988}44.6  & \cellcolor[rgb]{ .965,  .969,  .988}86.1  & \cellcolor[rgb]{ .965,  .969,  .988}23.1  & \cellcolor[rgb]{ .929,  .949,  .976}24.8  & \cellcolor[rgb]{ .929,  .949,  .976}27.2  & \cellcolor[rgb]{ .929,  .949,  .976}26.0  & \cellcolor[rgb]{ .929,  .949,  .976}29.0  & \cellcolor[rgb]{ .886,  .91,  .965}41.1  & \cellcolor[rgb]{ .886,  .91,  .965}45.3  & 39.3  & 62.3  & 0.130  \\
          & DiffAdvMAP & \cellcolor[rgb]{ .965,  .969,  .988}42.5  & \cellcolor[rgb]{ .965,  .969,  .988}47.2  & \cellcolor[rgb]{ .965,  .969,  .988}50.8  & \cellcolor[rgb]{ .965,  .969,  .988}99.9  & \cellcolor[rgb]{ .965,  .969,  .988}27.2  & \cellcolor[rgb]{ .929,  .949,  .976}29.7  & \cellcolor[rgb]{ .929,  .949,  .976}30.8  & \cellcolor[rgb]{ .929,  .949,  .976}27.9  & \cellcolor[rgb]{ .929,  .949,  .976}29.7  & \cellcolor[rgb]{ .886,  .91,  .965}36.9  & \cellcolor[rgb]{ .886,  .91,  .965}40.9  & 42.1  & 63.2  & 0.137  \\
          & IDATA(Our) & \cellcolor[rgb]{ .965,  .969,  .988}42.6 & \cellcolor[rgb]{ .965,  .969,  .988}43.2 & \cellcolor[rgb]{ .965,  .969,  .988}53.1 & \cellcolor[rgb]{ .965,  .969,  .988}97.2 & \cellcolor[rgb]{ .965,  .969,  .988}25.6 & \cellcolor[rgb]{ .929,  .949,  .976}31.2 & \cellcolor[rgb]{ .929,  .949,  .976}28.1 & \cellcolor[rgb]{ .929,  .949,  .976}28.8 & \cellcolor[rgb]{ .929,  .949,  .976}31.6 & \cellcolor[rgb]{ .886,  .91,  .965}43.1 & \cellcolor[rgb]{ .886,  .91,  .965}48.7 & \textcolor[rgb]{ 1,  0,  0}{\textbf{43.0 }} & \textcolor[rgb]{ 1,  0,  0}{\textbf{61.8 }} & \textcolor[rgb]{ 1,  0,  0}{\textbf{0.129 }} \\
    \midrule
    \multirow{9}[2]{*}{Swin-B} & DI-FGSM & \cellcolor[rgb]{ .965,  .969,  .988}47.3  & \cellcolor[rgb]{ .965,  .969,  .988}57.0  & \cellcolor[rgb]{ .965,  .969,  .988}58.5  & \cellcolor[rgb]{ .965,  .969,  .988}43.6  & \cellcolor[rgb]{ .965,  .969,  .988}36.1  & \cellcolor[rgb]{ .929,  .949,  .976}33.4  & \cellcolor[rgb]{ .929,  .949,  .976}97.3  & \cellcolor[rgb]{ .929,  .949,  .976}42.8  & \cellcolor[rgb]{ .929,  .949,  .976}42.0  & \cellcolor[rgb]{ .886,  .91,  .965}47.6  & \cellcolor[rgb]{ .886,  .91,  .965}49.2  & 45.8  & 65.7  & 0.138  \\
          & PI-FGSM & \cellcolor[rgb]{ .965,  .969,  .988}61.1  & \cellcolor[rgb]{ .965,  .969,  .988}58.3  & \cellcolor[rgb]{ .965,  .969,  .988}54.2  & \cellcolor[rgb]{ .965,  .969,  .988}64.3  & \cellcolor[rgb]{ .965,  .969,  .988}43.2  & \cellcolor[rgb]{ .929,  .949,  .976}51.2  & \cellcolor[rgb]{ .929,  .949,  .976}100  & \cellcolor[rgb]{ .929,  .949,  .976}47.6  & \cellcolor[rgb]{ .929,  .949,  .976}53.0  & \cellcolor[rgb]{ .886,  .91,  .965}56.5  & \cellcolor[rgb]{ .886,  .91,  .965}61.5  & 55.1  & 89.7  & 0.358  \\
          & ReColorAdv & \cellcolor[rgb]{ .965,  .969,  .988}36.5  & \cellcolor[rgb]{ .965,  .969,  .988}38.3  & \cellcolor[rgb]{ .965,  .969,  .988}42.1  & \cellcolor[rgb]{ .965,  .969,  .988}37.6  & \cellcolor[rgb]{ .965,  .969,  .988}37.2  & \cellcolor[rgb]{ .929,  .949,  .976}35.2  & \cellcolor[rgb]{ .929,  .949,  .976}100  & \cellcolor[rgb]{ .929,  .949,  .976}37.3  & \cellcolor[rgb]{ .929,  .949,  .976}35.4  & \cellcolor[rgb]{ .886,  .91,  .965}39.2  & \cellcolor[rgb]{ .886,  .91,  .965}44.8  & 38.4  & 65.7  & 0.147  \\
          & cAdv  & \cellcolor[rgb]{ .965,  .969,  .988}40.9  & \cellcolor[rgb]{ .965,  .969,  .988}40.9  & \cellcolor[rgb]{ .965,  .969,  .988}46.1  & \cellcolor[rgb]{ .965,  .969,  .988}43.2  & \cellcolor[rgb]{ .965,  .969,  .988}37.6  & \cellcolor[rgb]{ .929,  .949,  .976}51.7  & \cellcolor[rgb]{ .929,  .949,  .976}98.4  & \cellcolor[rgb]{ .929,  .949,  .976}33.4  & \cellcolor[rgb]{ .929,  .949,  .976}53.4  & \cellcolor[rgb]{ .886,  .91,  .965}54.0  & \cellcolor[rgb]{ .886,  .91,  .965}58.9  & 46.0  & 67.4  & 0.191  \\
          & VENOM & \cellcolor[rgb]{ .965,  .969,  .988}52.5  & \cellcolor[rgb]{ .965,  .969,  .988}55.9  & \cellcolor[rgb]{ .965,  .969,  .988}55.1  & \cellcolor[rgb]{ .965,  .969,  .988}61.0  & \cellcolor[rgb]{ .965,  .969,  .988}41.1  & \cellcolor[rgb]{ .929,  .949,  .976}42.7  & \cellcolor[rgb]{ .929,  .949,  .976}95.9  & \cellcolor[rgb]{ .929,  .949,  .976}45.7  & \cellcolor[rgb]{ .929,  .949,  .976}45.6  & \cellcolor[rgb]{ .886,  .91,  .965}55.4  & \cellcolor[rgb]{ .886,  .91,  .965}59.6  & 51.5  & 65.3  & 0.174  \\
          & ACA   & \cellcolor[rgb]{ .965,  .969,  .988}62.0  & \cellcolor[rgb]{ .965,  .969,  .988}64.2  & \cellcolor[rgb]{ .965,  .969,  .988}65.5  & \cellcolor[rgb]{ .965,  .969,  .988}66.1  & \cellcolor[rgb]{ .965,  .969,  .988}58.7  & \cellcolor[rgb]{ .929,  .949,  .976}59.1  & \cellcolor[rgb]{ .929,  .949,  .976}80.3  & \cellcolor[rgb]{ .929,  .949,  .976}62.1  & \cellcolor[rgb]{ .929,  .949,  .976}61.7  & \cellcolor[rgb]{ .886,  .91,  .965}70.9  & \cellcolor[rgb]{ .886,  .91,  .965}71.9  & 64.2  & 77.2  & 0.378  \\
          & DiffAttack & \cellcolor[rgb]{ .965,  .969,  .988}56.5  & \cellcolor[rgb]{ .965,  .969,  .988}57.9  & \cellcolor[rgb]{ .965,  .969,  .988}59.3  & \cellcolor[rgb]{ .965,  .969,  .988}58.6  & \cellcolor[rgb]{ .965,  .969,  .988}66.0  & \cellcolor[rgb]{ .929,  .949,  .976}61.0  & \cellcolor[rgb]{ .929,  .949,  .976}90.1  & \cellcolor[rgb]{ .929,  .949,  .976}65.0  & \cellcolor[rgb]{ .929,  .949,  .976}63.0  & \cellcolor[rgb]{ .886,  .91,  .965}62.3  & \cellcolor[rgb]{ .886,  .91,  .965}62.6  & 61.2  & 65.5  & 0.138  \\
          & DiffAdvMAP & \cellcolor[rgb]{ .965,  .969,  .988}58.8  & \cellcolor[rgb]{ .965,  .969,  .988}62.9  & \cellcolor[rgb]{ .965,  .969,  .988}65.5  & \cellcolor[rgb]{ .965,  .969,  .988}57.1  & \cellcolor[rgb]{ .965,  .969,  .988}75.7  & \cellcolor[rgb]{ .929,  .949,  .976}55.1  & \cellcolor[rgb]{ .929,  .949,  .976}95.9  & \cellcolor[rgb]{ .929,  .949,  .976}65.5  & \cellcolor[rgb]{ .929,  .949,  .976}59.8  & \cellcolor[rgb]{ .886,  .91,  .965}61.1  & \cellcolor[rgb]{ .886,  .91,  .965}60.9  & 62.2  & 66.4  & 0.131  \\
          & IDATA(Our) & \cellcolor[rgb]{ .965,  .969,  .988}63.5  & \cellcolor[rgb]{ .965,  .969,  .988}60.6  & \cellcolor[rgb]{ .965,  .969,  .988}65.3  & \cellcolor[rgb]{ .965,  .969,  .988}64.5  & \cellcolor[rgb]{ .965,  .969,  .988}60.0  & \cellcolor[rgb]{ .929,  .949,  .976}63.4  & \cellcolor[rgb]{ .929,  .949,  .976}96.3  & \cellcolor[rgb]{ .929,  .949,  .976}68.4  & \cellcolor[rgb]{ .929,  .949,  .976}66.7  & \cellcolor[rgb]{ .886,  .91,  .965}69.6  & \cellcolor[rgb]{ .886,  .91,  .965}68.0  & \textcolor[rgb]{ 1,  0,  0}{\textbf{65.0 }} & \textcolor[rgb]{ 1,  0,  0}{\textbf{63.8 }} & \textcolor[rgb]{ 1,  0,  0}{\textbf{0.128 }} \\
    \midrule
    \multirow{9}[2]{*}{Mix-B} & DI-FGSM & \cellcolor[rgb]{ .965,  .969,  .988}25.1  & \cellcolor[rgb]{ .965,  .969,  .988}32.5  & \cellcolor[rgb]{ .965,  .969,  .988}37.0  & \cellcolor[rgb]{ .965,  .969,  .988}38.3  & \cellcolor[rgb]{ .965,  .969,  .988}14.0  & \cellcolor[rgb]{ .929,  .949,  .976}26.5  & \cellcolor[rgb]{ .929,  .949,  .976}18.7  & \cellcolor[rgb]{ .929,  .949,  .976}27.3  & \cellcolor[rgb]{ .929,  .949,  .976}27.5  & \cellcolor[rgb]{ .886,  .91,  .965}100  & \cellcolor[rgb]{ .886,  .91,  .965}62.1  & 30.9  & 67.8  & 0.158  \\
          & PI-FGSM & \cellcolor[rgb]{ .965,  .969,  .988}23.1  & \cellcolor[rgb]{ .965,  .969,  .988}27.1  & \cellcolor[rgb]{ .965,  .969,  .988}29.7  & \cellcolor[rgb]{ .965,  .969,  .988}37.8  & \cellcolor[rgb]{ .965,  .969,  .988}6.7  & \cellcolor[rgb]{ .929,  .949,  .976}23.4  & \cellcolor[rgb]{ .929,  .949,  .976}9.1  & \cellcolor[rgb]{ .929,  .949,  .976}25.0  & \cellcolor[rgb]{ .929,  .949,  .976}22.7  & \cellcolor[rgb]{ .886,  .91,  .965}100  & \cellcolor[rgb]{ .886,  .91,  .965}49.3  & 25.4  & 87.8  & 0.147  \\
          & ReColorAdv & \cellcolor[rgb]{ .965,  .969,  .988}25.8  & \cellcolor[rgb]{ .965,  .969,  .988}28.6  & \cellcolor[rgb]{ .965,  .969,  .988}34.7  & \cellcolor[rgb]{ .965,  .969,  .988}38.6  & \cellcolor[rgb]{ .965,  .969,  .988}17.6  & \cellcolor[rgb]{ .929,  .949,  .976}27.1  & \cellcolor[rgb]{ .929,  .949,  .976}22.1  & \cellcolor[rgb]{ .929,  .949,  .976}25.5  & \cellcolor[rgb]{ .929,  .949,  .976}28.7  & \cellcolor[rgb]{ .886,  .91,  .965}97.8  & \cellcolor[rgb]{ .886,  .91,  .965}64.1  & 31.3  & 70.0  & 0.197  \\
          & cAdv  & \cellcolor[rgb]{ .965,  .969,  .988}24.5  & \cellcolor[rgb]{ .965,  .969,  .988}29.3  & \cellcolor[rgb]{ .965,  .969,  .988}30.0  & \cellcolor[rgb]{ .965,  .969,  .988}35.9  & \cellcolor[rgb]{ .965,  .969,  .988}12.4  & \cellcolor[rgb]{ .929,  .949,  .976}23.5  & \cellcolor[rgb]{ .929,  .949,  .976}16.3  & \cellcolor[rgb]{ .929,  .949,  .976}21.1  & \cellcolor[rgb]{ .929,  .949,  .976}24.7  & \cellcolor[rgb]{ .886,  .91,  .965}100  & \cellcolor[rgb]{ .886,  .91,  .965}55.2  & 27.3  & 66.9  & 0.204  \\
          & VENOM & \cellcolor[rgb]{ .965,  .969,  .988}51.5  & \cellcolor[rgb]{ .965,  .969,  .988}52.2  & \cellcolor[rgb]{ .965,  .969,  .988}55.9  & \cellcolor[rgb]{ .965,  .969,  .988}57.0  & \cellcolor[rgb]{ .965,  .969,  .988}40.9  & \cellcolor[rgb]{ .929,  .949,  .976}44.0  & \cellcolor[rgb]{ .929,  .949,  .976}49.8  & \cellcolor[rgb]{ .929,  .949,  .976}46.4  & \cellcolor[rgb]{ .929,  .949,  .976}49.3  & \cellcolor[rgb]{ .886,  .91,  .965}82.2  & \cellcolor[rgb]{ .886,  .91,  .965}73.5  & 52.1  & 77.6  & 0.207  \\
          & ACA   & \cellcolor[rgb]{ .965,  .969,  .988}52.2  & \cellcolor[rgb]{ .965,  .969,  .988}57.6  & \cellcolor[rgb]{ .965,  .969,  .988}60.4  & \cellcolor[rgb]{ .965,  .969,  .988}58.6  & \cellcolor[rgb]{ .965,  .969,  .988}45.5  & \cellcolor[rgb]{ .929,  .949,  .976}54.8  & \cellcolor[rgb]{ .929,  .949,  .976}49.9  & \cellcolor[rgb]{ .929,  .949,  .976}51.9  & \cellcolor[rgb]{ .929,  .949,  .976}52.9  & \cellcolor[rgb]{ .886,  .91,  .965}92.5  & \cellcolor[rgb]{ .886,  .91,  .965}68.4  & 55.2  & 85.3  & 0.386  \\
          & DiffAttack & \cellcolor[rgb]{ .965,  .969,  .988}48.0  & \cellcolor[rgb]{ .965,  .969,  .988}47.4  & \cellcolor[rgb]{ .965,  .969,  .988}51.5  & \cellcolor[rgb]{ .965,  .969,  .988}55.6  & \cellcolor[rgb]{ .965,  .969,  .988}42.9  & \cellcolor[rgb]{ .929,  .949,  .976}50.4  & \cellcolor[rgb]{ .929,  .949,  .976}49.7  & \cellcolor[rgb]{ .929,  .949,  .976}54.9  & \cellcolor[rgb]{ .929,  .949,  .976}53.2  & \cellcolor[rgb]{ .886,  .91,  .965}83.1  & \cellcolor[rgb]{ .886,  .91,  .965}77.7  & 53.1  & 65.3  & 0.139  \\
          & DiffAdvMAP & \cellcolor[rgb]{ .965,  .969,  .988}44.1  & \cellcolor[rgb]{ .965,  .969,  .988}48.2  & \cellcolor[rgb]{ .965,  .969,  .988}52.1  & \cellcolor[rgb]{ .965,  .969,  .988}49.3  & \cellcolor[rgb]{ .965,  .969,  .988}37.6  & \cellcolor[rgb]{ .929,  .949,  .976}47.5  & \cellcolor[rgb]{ .929,  .949,  .976}42.3  & \cellcolor[rgb]{ .929,  .949,  .976}48.7  & \cellcolor[rgb]{ .929,  .949,  .976}51.7  & \cellcolor[rgb]{ .886,  .91,  .965}98.9  & \cellcolor[rgb]{ .886,  .91,  .965}85.4  & 50.7  & 66.4  & 0.139  \\
          & IDATA(Our) & \cellcolor[rgb]{ .965,  .969,  .988}53.0  & \cellcolor[rgb]{ .965,  .969,  .988}51.5  & \cellcolor[rgb]{ .965,  .969,  .988}55.9  & \cellcolor[rgb]{ .965,  .969,  .988}60.6  & \cellcolor[rgb]{ .965,  .969,  .988}43.7  & \cellcolor[rgb]{ .929,  .949,  .976}58.9  & \cellcolor[rgb]{ .929,  .949,  .976}45.7  & \cellcolor[rgb]{ .929,  .949,  .976}56.8  & \cellcolor[rgb]{ .929,  .949,  .976}54.2  & \cellcolor[rgb]{ .886,  .91,  .965}96.4  & \cellcolor[rgb]{ .886,  .91,  .965}85.5  & \textcolor[rgb]{ 1,  0,  0}{\textbf{56.6 }} & \textcolor[rgb]{ 1,  0,  0}{\textbf{64.3 }} & \textcolor[rgb]{ 1,  0,  0}{\textbf{0.138 }} \\
    \bottomrule
    \end{tabular}
    }
  \label{tab:tabble1}
\end{table*}

\section{Experiment}
\subsection{Experimental Setup}
\textbf{Datasets:} 
Following prior methods \cite{DiffAttack, kuurila2025venom, Chen2023ContentbasedUA, pan2024sca}, we evaluate the performance of IDATA on the ImageNet-Compatible Dataset, a standard benchmark for transferability evaluation. 
This dataset comprises 1,000 images and each evaluation sample is resized to $224 \times 224$. We also conduct further experiments on the CUB-200-2011 Dataset~\cite{wah2011caltech} and the
Stanford Cars Dataset~\cite{krause20133d}.

\noindent \textbf{Target Models and Defenses:}
We evaluate both transferability and visual imperceptibility on diverse architectures spanning CNNs, Transformers, and MLPs.
The CNN-based models include ConvNeXt~\cite{liu2022convnet}, ResNet-50 (Res-50)~\cite{he2016deep}, VGG-19~\cite{simonyan2014very}, Inception-v3 (Inc-v3)~\cite{szegedy2016rethinking}, and MobileNet-v2 (Mob-v2)~\cite{sandler2018mobilenetv2}.  
The Transformer models include ViT-B/16 (ViT-B)~\cite{dosovitskiy2021an}, Swin-B~\cite{liu2021swin}, DeiT-B, and DeiT-S~\cite{touvron2021training}. 
The MLP-based models include Mixer-B/16 (Mix-B) and Mixer-L/16 (Mix-L)~\cite{tolstikhin2021mlp}.
To evaluate robustness under defended settings, we further consider representative purification-based defenses, including NRP~\cite{naseer2020self}, DiffPure~\cite{nie2022diffusion}, OSGD~\cite{li2025adversarial}, GuidedDiffusionPur~\cite{wang2022guided}, as well as adversarially trained models, including Adv-Inc-v3~\cite{kurakin2018adversarial}, Inc-v3ens3, IncRes-v2ens~\cite{tramer2017ensemble} and Inc-v3ens4.

\noindent \textbf{Implementation Details:}
We adopt DDIM sampling on Stable Diffusion v2.0~\cite{rombach2022high} with $T=50$ diffusion steps. The diffusion process is conditioned on the ground-truth label name of $y$. The guidance scale is $0$ during inversion and $1$ during reverse denoising. For adversarial generation, the latent at step $\tau=20$ is used to initialize reverse denoising. The unified adversarial latent perturbation $\bm{\delta}$ is optimized with momentum factor $\mu=1$, step size $\eta=0.01$, loss weight $\lambda=2$, and $20$ iterations.

\begin{table*}[t]
  \renewcommand{\arraystretch}{0.5}
  \centering
  \caption{\textbf{Robustness against defenses.} ``AVG'' denotes the average ASR (\%) over all defended models. ``Average (w/o self)'' denotes the average ASR drop across all defenses excluding the $\text{mixer-b}_{\text{normal}}$; higher values indicate better robustness. The best and second-best results are highlighted in dark and light blue, respectively.}
  \resizebox{0.98\textwidth}{!}{
    \begin{tabular}{cccccc|cccccc}
    \toprule
    \multirow{2}[4]{*}{Attacks} & \multicolumn{5}{c|}{Defended Models}   & \multicolumn{6}{c}{Defended Methods} \\
\cmidrule{2-12}          & \multicolumn{1}{c}{\multirow{1}[2]{*}{Adv-inc-v3}} & \multicolumn{1}{c}{\multirow{1}[2]{*}{Inc-v3ens3}} & \multicolumn{1}{c}{\multirow{1}[2]{*}{Inc-v3ens4}} & \multirow{1}[2]{*}{IncRes-v2ens} & \multirow{1}[2]{*}{AVG $\uparrow$} & \multirow{1}[2]{*}{$\text{mixer-b}_{\text{normal}}$} & \multicolumn{1}{c}{\multirow{1}[2]{*}{NRP}} & \multicolumn{1}{c}{\multirow{1}[2]{*}{OSGD}} & \multicolumn{1}{c}{\multirow{1}[2]{*}{DiffPure}} & \multicolumn{1}{c}{\multirow{1}[2]{*}{GuidedDiffusionPur}} & \multicolumn{1}{c}{\multirow{1}[2]{*}{Average\newline{}(w/o self)$\uparrow$}} \\
          &       &       &       &       &       &       &       &       &       &       &  \\
    \midrule
    \midrule
    DI-FGSM & 31.7  & 34.7  & 31    & 20.5  & 29.48  & \cellcolor[rgb]{ .71,  .776,  .918}100  & -40.0  & -48.5  & -52.4  & -59.5  & -50.1  \\
    PI-FGSM & 20.8  & 25.3  & 25.5  & 12.2  & 20.95  & \cellcolor[rgb]{ .71,  .776,  .918}100  & -42.4  & -42.1  & -41.5  & -55.5  & -45.4  \\
    ReColorAdv & 24.6  & 30.7  & 30.5  & 19.5  & 26.33  & 97.8  & -16.5  & -39.4  & -47.8  & -48.7  & -38.1  \\
    cAdv  & 31.5  & 34.6  & 36.3  & 24.5  & 31.73  & \cellcolor[rgb]{ .71,  .776,  .918}100  & -23.3  & -33.4  & -39.2  & -40.2  & -34.0  \\
    VENOM & 59.3  & 59.5  & 62.1  & 50.3  & 57.80  & 82.2  & -13.3  & -16.5  & -13.4  & -29.8  & -18.3  \\
    ACA   & \cellcolor[rgb]{ .91,  .929,  .973}60.6 & \cellcolor[rgb]{ .91,  .929,  .973}63.8 & \cellcolor[rgb]{ .91,  .929,  .973}62.5 & \cellcolor[rgb]{ .71,  .776,  .918}\textbf{56.9} & \cellcolor[rgb]{ .91,  .929,  .973}60.95  & 92.5  & -12.4  & -13.5  & -15.8  & -21.1  & -15.7  \\
    DiffAttack & 48.5  & 49.9  & 51.8  & 39.8  & 47.50  & 83.1  & -16.6  & -15.2  & -13.9  & -26.5  & -18.1  \\
    DiffAdvMAP & 41.9  & 47.1  & 45    & 34.3  & 42.08  & \cellcolor[rgb]{ .91,  .929,  .973}98.9  & \cellcolor[rgb]{ .91,  .929,  .973}-10.9  & \cellcolor[rgb]{ .91,  .929,  .973}-10.5  & \cellcolor[rgb]{ .91,  .929,  .973}-12.4  & \cellcolor[rgb]{ .91,  .929,  .973}-20.1  & \cellcolor[rgb]{ .91,  .929,  .973}-13.5  \\
    IDATA(Our) & \cellcolor[rgb]{ .71,  .776,  .918}\textbf{64.4} & \cellcolor[rgb]{ .71,  .776,  .918}\textbf{65.6} & \cellcolor[rgb]{ .71,  .776,  .918}\textbf{64.6} & \cellcolor[rgb]{ .91,  .929,  .973}56.5 & \cellcolor[rgb]{ .71,  .776,  .918}\textbf{62.78}  & 96.4  & \cellcolor[rgb]{ .71,  .776,  .918}\textbf{-4.9 } & \cellcolor[rgb]{ .71,  .776,  .918}\textbf{-5.2 } & \cellcolor[rgb]{ .71,  .776,  .918}\textbf{-6.8 } & \cellcolor[rgb]{ .71,  .776,  .918}\textbf{-8.0 } & \cellcolor[rgb]{ .71,  .776,  .918}\textbf{-6.2}  \\
    \bottomrule
    \end{tabular}
    }
  \label{tab:table2}
\end{table*}

\begin{figure*}[t]
  \centering
  \includegraphics[width=0.98\linewidth]{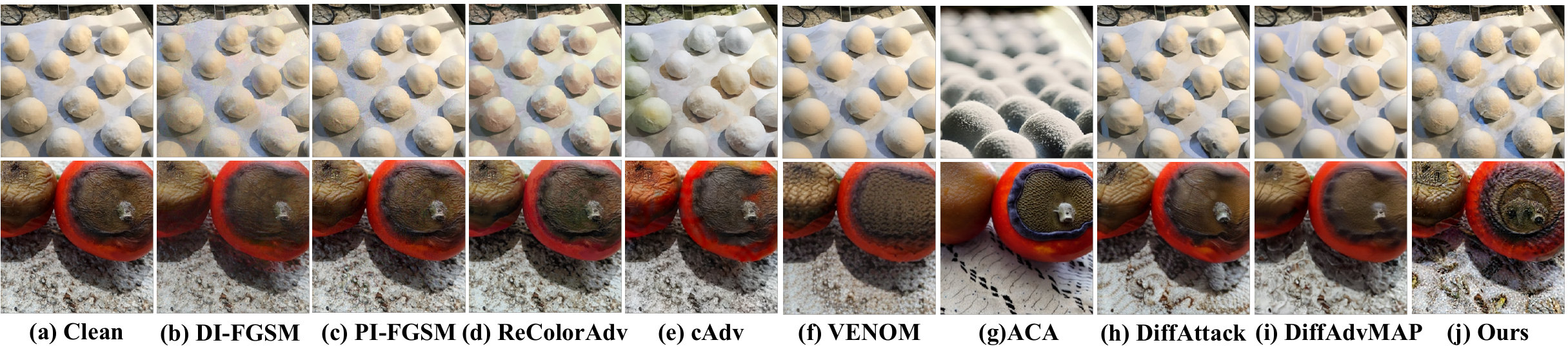}
    \caption{Visual comparisons among different adversarial attacks. Please zoom in for a better view.}
    \label{pic:visualization}
\end{figure*}

\noindent \textbf{Comparison Methods:}
We compare IDATA with representative transfer attacks from three categories: pixel-space transfer attacks DI-FGSM~\cite{xie2019improving} and PI-FGSM~\cite{gao2020patch}, unrestricted transfer attacks ReColorAdv~\cite{NEURIPS2019_6e923226} and cAdv~\cite{bhattad2019unrestricted}, and diffusion-based attacks DiffAttack~\cite{DiffAttack}, VENOM~\cite{kuurila2025venom}, ACA~\cite{Chen2023ContentbasedUA}, and DiffAdvMAP~\cite{pan2025diffadvmap}. All baselines follow their official or default optimal settings for fair comparison.

\noindent \textbf{Evaluation Metrics:}
We evaluate transferability via the Average Attack Success Rate (AVG), where higher values indicate better cross-model performance.
For visual imperceptibility, we use Learned Perceptual Image Patch Similarity (LPIPS)~\cite{zhang2018unreasonable} and Fréchet Inception Distance (FID)~\cite{heusel2017gans}. LPIPS quantifies local artifacts and FID captures global semantic consistency, with lower values indicating higher perceptual similarity and better imperceptibility.

\subsection{Evaluation Results}
\textit{1) Results on Normally Trained Models:}
Tab.~\ref{tab:tabble1} summarizes results on normally trained models. Overall, IDATA achieves the optimal trade-off between black-box transferability and visual imperceptibility across all surrogate settings.

\textbf{Superior Transferability and Robustness.} 
IDATA consistently achieves the highest AVG across the four diverse surrogate models. Notably, on the challenging Mix-B surrogate, IDATA outperforms state-of-the-art diffusion-based baselines, DiffAttack~\cite{DiffAttack} and DiffAdvMAP~\cite{pan2025diffadvmap}, by 3.5\% and 5.9\%, respectively. While methods like VENOM~\cite{kuurila2025venom} and ACA~\cite{Chen2023ContentbasedUA} occasionally yield competitive scores on specific targets, Tab.~\ref{tab:tabble1} shows they severely compromise visual quality to do so. Furthermore, instead of overfitting to surrogate decision boundaries, IDATA exhibits remarkable cross-architecture generalization, maintaining top-tier performance across CNNs, Transformers, and MLPs. We attribute this stable robustness to IDM's deep trajectory optimization for universal semantics, combined with LFCM's ability to steer perturbations toward model-agnostic, low-frequency structures.

\begin{figure}[t]
    \centering
    \includegraphics[width=\linewidth]{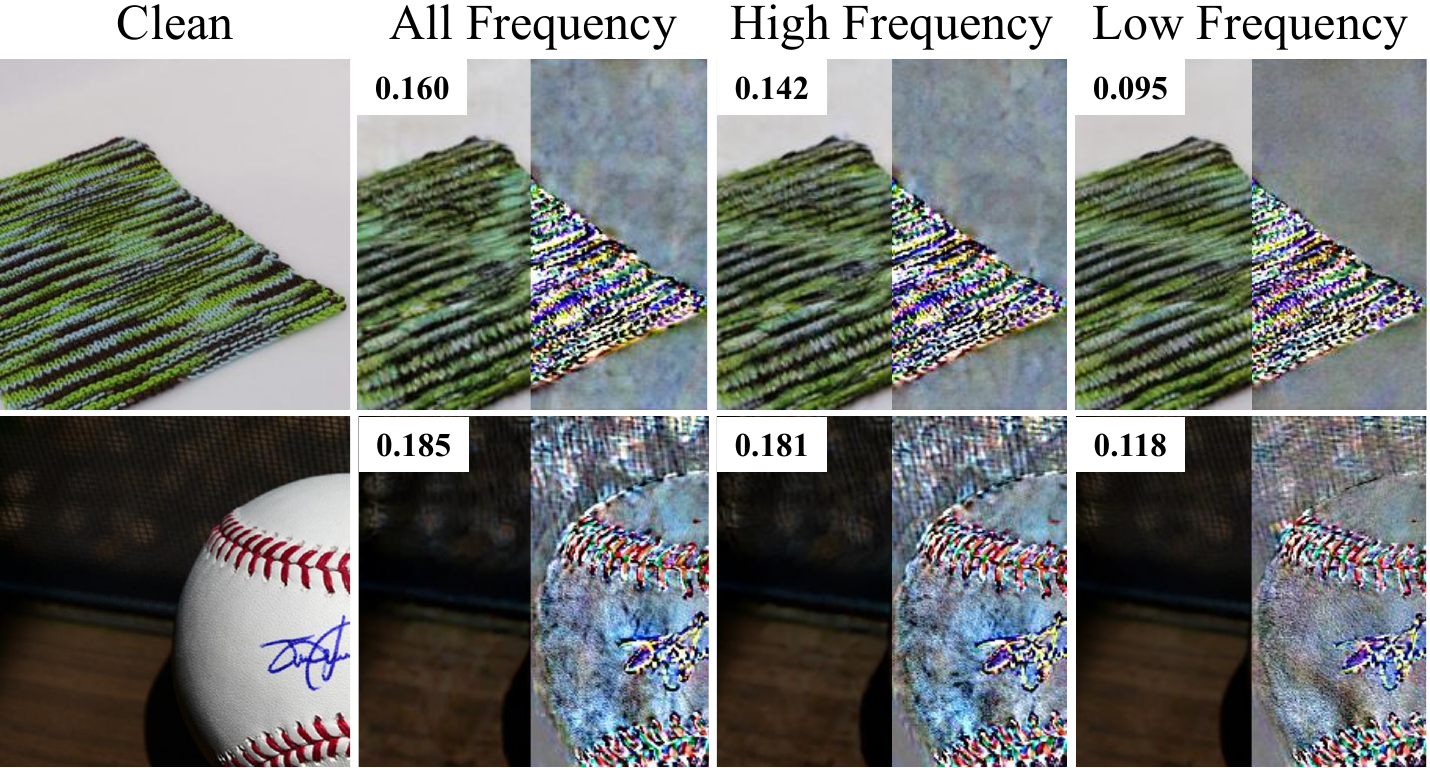}
    \caption{Comparison of adversarial examples generated by frequency-specific perturbations in the latent space. For each example, the left half shows the adversarial image, and the right half shows the color-coded perturbation map, where gray denotes no perturbation, brighter colors denote positive perturbations, and darker colors denote negative perturbations. LPIPS ($\downarrow$) is reported in the top left.}
    \label{pic:ablation_vis}
\end{figure}

\textbf{Exceptional Visual Imperceptibility.}
Beyond its transferability, IDATA establishes a new benchmark for visual imperceptibility.
This is first quantified by its consistently superior FID and LPIPS scores across all surrogate settings, as shown in Tab.~\ref{tab:tabble1}.
Notably, our method strictly maintains the LPIPS distortion below 0.14, a tighter budget than most baselines.

The qualitative comparisons in Fig.~\ref{pic:visualization} further support these observations. Adversarial examples generated by IDATA remain visually close to the clean images, without noticeable artifacts, color shifts, or over-sharpening. In contrast, pixel-based attacks such as DI-FGSM~\cite{xie2019improving} and PI-FGSM~\cite{gao2020patch} introduce visible high-frequency noise, while unrestricted attacks such as ReColorAdv~\cite{NEURIPS2019_6e923226} and cAdv~\cite{bhattad2019unrestricted} often distort image colors. Although diffusion-based baselines, including DiffAttack~\cite{DiffAttack}, ACA~\cite{Chen2023ContentbasedUA}, DiffAdvMAP~\cite{pan2025diffadvmap}, and VENOM~\cite{kuurila2025venom}, yield more natural results but still exhibit over-sharpening artifacts. For instance, the ``steamed bun'' in the first column of Fig.~\ref{pic:visualization} appears unnaturally sharpened in their results.

By comparison, IDATA better preserves natural textures and object structures while maintaining strong attack transferability. This advantage stems from deep adversarial optimization over invertible diffusion trajectories, together with low-frequency regularization and perceptual guidance, which encourage perturbations to reside in semantically stable latent components. As a result, IDATA achieves a clearly more favorable transferability-imperceptibility trade-off than existing baselines.

\begin{table*}[t]
  \renewcommand{\arraystretch}{0.5}
  \centering
  \caption{Comparison on CUB-200-2011 and Stanford Cars datasets. The best results are highlighted in bold \textcolor{red}{red} font.}
  \resizebox{0.98\textwidth}{!}{
        \begin{tabular}{cc|ccc|ccc|ccc|ccc}
    \toprule
    \multirow{3}[6]{*}{\diagbox[width=2em]{S.}{T.}} & \multirow{2}[4]{*}{Attacks} & \multicolumn{6}{c|}{CUB-200-2011}             & \multicolumn{6}{c}{Stanford Cars} \\
\cmidrule{3-14}          &       & Res-50 & SENET154 & SE-Res101 & \multicolumn{1}{p{6.32em}}{AVG (w/o self)$\uparrow$} & FID$\downarrow$   & LPIPS$\downarrow$ & Res-50 & SENET154 & SE-Res101 & \multicolumn{1}{p{6.32em}}{AVG (w/o self)$\uparrow$} & FID$\downarrow$   & LPIPS$\downarrow$ \\
\cmidrule{2-14}          & Clean & 24.3  & 19.5  & 23.4  & 22.4  & 11.1  & -     & 26.1  & 23.6  & 25.6  & \multicolumn{1}{c|}{25.1} & 11.6  & - \\
    \midrule
    \midrule
    \multirow{8}[2]{*}{Res-50} & DI-FGSM & \cellcolor[rgb]{ .965,  .969,  .988}99.7  & \cellcolor[rgb]{ .965,  .969,  .988}57.3  & \cellcolor[rgb]{ .965,  .969,  .988}66.2  & 61.8  & 20.9  & 0.155  & \cellcolor[rgb]{ .965,  .969,  .988}99.9  & \cellcolor[rgb]{ .965,  .969,  .988}66.7  & \cellcolor[rgb]{ .965,  .969,  .988}70.7  & 68.7  & 28.7  & 0.097  \\
          & PI-FGSM & \cellcolor[rgb]{ .965,  .969,  .988}90.9  & \cellcolor[rgb]{ .965,  .969,  .988}64.8  & \cellcolor[rgb]{ .965,  .969,  .988}73.8  & 69.3  & 34.8  & 0.355  & \cellcolor[rgb]{ .965,  .969,  .988}98.5  & \cellcolor[rgb]{ .965,  .969,  .988}68.5  & \cellcolor[rgb]{ .965,  .969,  .988}76.8  & 72.7  & 53.2  & 0.310  \\
          & ReColorAdv & \cellcolor[rgb]{ .965,  .969,  .988}99.9  & \cellcolor[rgb]{ .965,  .969,  .988}58.0  & \cellcolor[rgb]{ .965,  .969,  .988}66.6  & 62.3  & 23.2  & 0.215  & \cellcolor[rgb]{ .965,  .969,  .988}100  & \cellcolor[rgb]{ .965,  .969,  .988}57.4  & \cellcolor[rgb]{ .965,  .969,  .988}64.9  & 61.2  & 22.9  & 0.164  \\
          & cAdv  & \cellcolor[rgb]{ .965,  .969,  .988}75.0  & \cellcolor[rgb]{ .965,  .969,  .988}60.0  & \cellcolor[rgb]{ .965,  .969,  .988}63.7  & 61.9  & 21.3  & 0.129  & \cellcolor[rgb]{ .965,  .969,  .988}61.9  & \cellcolor[rgb]{ .965,  .969,  .988}35.3  & \cellcolor[rgb]{ .965,  .969,  .988}39.1  & 37.2  & 19.7  & 0.117  \\
          & VENOM & \cellcolor[rgb]{ .965,  .969,  .988}97.3  & \cellcolor[rgb]{ .965,  .969,  .988}67.5  & \cellcolor[rgb]{ .965,  .969,  .988}69.6  & 68.6  & 68.1  & 0.135  & \cellcolor[rgb]{ .965,  .969,  .988}92.9  & \cellcolor[rgb]{ .965,  .969,  .988}70.1  & \cellcolor[rgb]{ .965,  .969,  .988}71.2  & 70.7  & 26.7  & 0.135  \\
          & DiffAttack & \cellcolor[rgb]{ .965,  .969,  .988}96.7  & \cellcolor[rgb]{ .965,  .969,  .988}80.7  & \cellcolor[rgb]{ .965,  .969,  .988}83.3  & 82.0  & 20.6  & 0.122  & \cellcolor[rgb]{ .965,  .969,  .988}99.9  & \cellcolor[rgb]{ .965,  .969,  .988}84.9  & \cellcolor[rgb]{ .965,  .969,  .988}86.9  & 85.9  & 17.8  & 0.112  \\
          & DiffAdvMAP & \cellcolor[rgb]{ .965,  .969,  .988}99.6  & \cellcolor[rgb]{ .965,  .969,  .988}81.9  & \cellcolor[rgb]{ .965,  .969,  .988}83.9  & 82.9  & 24.0  & 0.116  & \cellcolor[rgb]{ .965,  .969,  .988}99.7  & \cellcolor[rgb]{ .965,  .969,  .988}83.7  & \cellcolor[rgb]{ .965,  .969,  .988}86.1  & 84.9  & 18.3  & 0.117  \\
          & IDATA(Our) & \cellcolor[rgb]{ .965,  .969,  .988}94.2 & \cellcolor[rgb]{ .965,  .969,  .988}83.3  & \cellcolor[rgb]{ .965,  .969,  .988}84.6  & \textcolor[rgb]{ 1,  0,  0}{\textbf{84.0 }} & \textcolor[rgb]{ 1,  0,  0}{\textbf{18.9 }} & \textcolor[rgb]{ 1,  0,  0}{\textbf{0.115 }} & \cellcolor[rgb]{ .965,  .969,  .988}96.4  & \cellcolor[rgb]{ .965,  .969,  .988}84.4  & \cellcolor[rgb]{ .965,  .969,  .988}88.1  & \textcolor[rgb]{ 1,  0,  0}{\textbf{86.3 }} & \textcolor[rgb]{ 1,  0,  0}{\textbf{15.3 }} & \textcolor[rgb]{ 1,  0,  0}{\textbf{0.110 }} \\
    \midrule
    \multirow{8}[2]{*}{SENET154} & DI-FGSM & \cellcolor[rgb]{ .965,  .969,  .988}45.5  & \cellcolor[rgb]{ .965,  .969,  .988}99.8  & \cellcolor[rgb]{ .965,  .969,  .988}51.1 & 48.3  & 23.5  & 0.158  & \cellcolor[rgb]{ .965,  .969,  .988}54.4  & \cellcolor[rgb]{ .965,  .969,  .988}99.9  & \cellcolor[rgb]{ .965,  .969,  .988}54.5  & 54.5  & 29.1  & 0.096  \\
          & PI-FGSM & \cellcolor[rgb]{ .965,  .969,  .988}39.9  & \cellcolor[rgb]{ .965,  .969,  .988}99.7  & \cellcolor[rgb]{ .965,  .969,  .988}43.8 & 41.9  & 20.8  & 0.137  & \cellcolor[rgb]{ .965,  .969,  .988}45.9  & \cellcolor[rgb]{ .965,  .969,  .988}99.9  & \cellcolor[rgb]{ .965,  .969,  .988}46.8  & 46.4  & 23.0  & 0.095  \\
          & ReColorAdv & \cellcolor[rgb]{ .965,  .969,  .988}44.8  & \cellcolor[rgb]{ .965,  .969,  .988}95.7  & \cellcolor[rgb]{ .965,  .969,  .988}51.1 & 48.0  & 22.4  & 0.153  & \cellcolor[rgb]{ .965,  .969,  .988}55.3  & \cellcolor[rgb]{ .965,  .969,  .988}100  & \cellcolor[rgb]{ .965,  .969,  .988}57.1  & 56.2  & 21.3  & 0.130  \\
          & cAdv  & \cellcolor[rgb]{ .965,  .969,  .988}69.0  & \cellcolor[rgb]{ .965,  .969,  .988}94.3  & \cellcolor[rgb]{ .965,  .969,  .988}68.7 & 68.9  & 20.4  & 0.118  & \cellcolor[rgb]{ .965,  .969,  .988}36.7  & \cellcolor[rgb]{ .965,  .969,  .988}79.8  & \cellcolor[rgb]{ .965,  .969,  .988}39.9  & 38.3  & 17.8  & 0.102  \\
          & VENOM & \cellcolor[rgb]{ .965,  .969,  .988}68.1  & \cellcolor[rgb]{ .965,  .969,  .988}93.7  & \cellcolor[rgb]{ .965,  .969,  .988}69.8 & 69.0  & 61.6  & 0.126  & \cellcolor[rgb]{ .965,  .969,  .988}59.9  & \cellcolor[rgb]{ .965,  .969,  .988}89.7  & \cellcolor[rgb]{ .965,  .969,  .988}60.4  & 60.2  & 16.2  & 0.086  \\
          & DiffAttack & \cellcolor[rgb]{ .965,  .969,  .988}46.2  & \cellcolor[rgb]{ .965,  .969,  .988}97.5  & \cellcolor[rgb]{ .965,  .969,  .988}48.7 & 47.5  & 17.9  & 0.104  & \cellcolor[rgb]{ .965,  .969,  .988}62.7  & \cellcolor[rgb]{ .965,  .969,  .988}99.1  & \cellcolor[rgb]{ .965,  .969,  .988}67.5  & 65.1  & 16.2  & \textcolor[rgb]{ 1,  0,  0}{\textbf{0.095 }} \\
          & DiffAdvMAP & \cellcolor[rgb]{ .965,  .969,  .988}65.5  & \cellcolor[rgb]{ .965,  .969,  .988}98.3  & \cellcolor[rgb]{ .965,  .969,  .988}69.8 & 67.7  & 20.0  & 0.108  & \cellcolor[rgb]{ .965,  .969,  .988}66.7  & \cellcolor[rgb]{ .965,  .969,  .988}100  & \cellcolor[rgb]{ .965,  .969,  .988}69.9  & 68.3  & 16.3  & 0.103  \\
          & IDATA(Our) & \cellcolor[rgb]{ .965,  .969,  .988}67.5  & \cellcolor[rgb]{ .965,  .969,  .988}95.0  & \cellcolor[rgb]{ .965,  .969,  .988}72.3  & \textcolor[rgb]{ 1,  0,  0}{\textbf{69.9 }} & \textcolor[rgb]{ 1,  0,  0}{\textbf{16.2 }} & \textcolor[rgb]{ 1,  0,  0}{\textbf{0.102 }} & \cellcolor[rgb]{ .965,  .969,  .988}69.8  & \cellcolor[rgb]{ .965,  .969,  .988}98.6  & \cellcolor[rgb]{ .965,  .969,  .988}73.1  & \textcolor[rgb]{ 1,  0,  0}{\textbf{71.5 }} & \textcolor[rgb]{ 1,  0,  0}{\textbf{14.6 }} & 0.097  \\
    \midrule
    \multirow{8}[2]{*}{SE-Res101} & DI-FGSM & \cellcolor[rgb]{ .965,  .969,  .988}60.6  & \cellcolor[rgb]{ .965,  .969,  .988}62.0  & \cellcolor[rgb]{ .965,  .969,  .988}99.8 & 61.3  & 23.5  & 0.165  & \cellcolor[rgb]{ .965,  .969,  .988}71.9  & \cellcolor[rgb]{ .965,  .969,  .988}70.7  & \cellcolor[rgb]{ .965,  .969,  .988}99.8  & 71.3  & 28.5  & 0.106  \\
          & PI-FGSM & \cellcolor[rgb]{ .965,  .969,  .988}78.3  & \cellcolor[rgb]{ .965,  .969,  .988}70.2  & \cellcolor[rgb]{ .965,  .969,  .988}100 & 74.3  & 45.5  & 0.403  & \cellcolor[rgb]{ .965,  .969,  .988}81.5  & \cellcolor[rgb]{ .965,  .969,  .988}70.7  & \cellcolor[rgb]{ .965,  .969,  .988}100  & 76.1  & 59.9  & 0.331  \\
          & ReColorAdv & \cellcolor[rgb]{ .965,  .969,  .988}58.4  & \cellcolor[rgb]{ .965,  .969,  .988}57.8  & \cellcolor[rgb]{ .965,  .969,  .988}99.4 & 58.1  & 27.0  & 0.198  & \cellcolor[rgb]{ .965,  .969,  .988}63.2  & \cellcolor[rgb]{ .965,  .969,  .988}58.5  & \cellcolor[rgb]{ .965,  .969,  .988}100  & 60.9  & 23.2  & 0.155  \\
          & cAdv  & \cellcolor[rgb]{ .965,  .969,  .988}66.9  & \cellcolor[rgb]{ .965,  .969,  .988}66.2  & \cellcolor[rgb]{ .965,  .969,  .988}89.5 & 66.6  & 22.7  & 0.125  & \cellcolor[rgb]{ .965,  .969,  .988}37.6  & \cellcolor[rgb]{ .965,  .969,  .988}37.3  & \cellcolor[rgb]{ .965,  .969,  .988}77.5  & 37.5  & 18.5  & 0.119  \\
          & VENOM & \cellcolor[rgb]{ .965,  .969,  .988}67.8  & \cellcolor[rgb]{ .965,  .969,  .988}68.9  & \cellcolor[rgb]{ .965,  .969,  .988}95.4 & 68.4  & 74.8  & 0.127  & \cellcolor[rgb]{ .965,  .969,  .988}69.9  & \cellcolor[rgb]{ .965,  .969,  .988}71.4  & \cellcolor[rgb]{ .965,  .969,  .988}92.2  & 70.7  & 20.0  & 0.122  \\
          & DiffAttack & \cellcolor[rgb]{ .965,  .969,  .988}73.0  & \cellcolor[rgb]{ .965,  .969,  .988}76.5  & \cellcolor[rgb]{ .965,  .969,  .988}96.1 & 74.8  & 22.4  & 0.121  & \cellcolor[rgb]{ .965,  .969,  .988}82.5  & \cellcolor[rgb]{ .965,  .969,  .988}84.0  & \cellcolor[rgb]{ .965,  .969,  .988}99.7  & 83.3  & 18.0  & 0.114  \\
          & DiffAdvMAP & \cellcolor[rgb]{ .965,  .969,  .988}82.9  & \cellcolor[rgb]{ .965,  .969,  .988}83.7  & \cellcolor[rgb]{ .965,  .969,  .988}99.9 & 83.3  & 22.1  & 0.115  & \cellcolor[rgb]{ .965,  .969,  .988}84.0  & \cellcolor[rgb]{ .965,  .969,  .988}85.8  & \cellcolor[rgb]{ .965,  .969,  .988}100  & 84.9  & 18.3  & 0.116  \\
          & IDATA(Our) & \cellcolor[rgb]{ .965,  .969,  .988}83.2  & \cellcolor[rgb]{ .965,  .969,  .988}84.1  & \cellcolor[rgb]{ .965,  .969,  .988}98.7 & \textcolor[rgb]{ 1,  0,  0}{\textbf{83.7 }} & \textcolor[rgb]{ 1,  0,  0}{\textbf{20.7 }} & \textcolor[rgb]{ 1,  0,  0}{\textbf{0.114 }} & \cellcolor[rgb]{ .965,  .969,  .988}85.0  & \cellcolor[rgb]{ .965,  .969,  .988}86.5  & \cellcolor[rgb]{ .965,  .969,  .988}99.2  & \textcolor[rgb]{ 1,  0,  0}{\textbf{85.8 }} & \textcolor[rgb]{ 1,  0,  0}{\textbf{17.3 }} & \textcolor[rgb]{ 1,  0,  0}{\textbf{0.113 }} \\
    \bottomrule
    \end{tabular}
    }
  \label{tab:table3}
\end{table*}

\textit{2) Robustness Against Defense Approaches:}
Using Mixer-B~\cite{tolstikhin2021mlp} as the surrogate, we evaluate IDATA against standard purification methods and adversarially trained models. Tab.~\ref{tab:table2} shows that IDATA consistently outperforms baselines across all defense settings. Crucially, it demonstrates exceptional resistance to image purification, with only a 6.8\% performance drop under DiffPure~\cite{nie2022diffusion} compared to 12.4\% or worse for other methods. IDATA also remains highly potent against adversarially trained models. This confirms that our method crafts highly transferable perturbations that are inherently robust against common defensive interventions.

\textit{3) Performance on More Datasets:}
As shown in Tab.~\ref{tab:table3}, IDATA achieves the best overall transferability on the CUB-200-2011 and Stanford Cars datasets. It consistently outperforms DiffAttack~\cite{DiffAttack} and DiffAdvMAP~\cite{pan2025diffadvmap} across different surrogate models. Critically, this is achieved while delivering superior visual quality; for instance, IDATA obtains the lowest LPIPS in all CUB-200-2011 settings and the best FID scores across all surrogate models on Stanford Cars.

\begin{table}[t]
  \renewcommand{\arraystretch}{0.5}
  \footnotesize
  \centering
  \caption{Effectiveness of IDM and LFCM of IDATA.}
  \resizebox{0.46\textwidth}{!}{
    \begin{tabular}{rc|ccc}
    \toprule
    \multicolumn{1}{c}{IDM} & LFCM  & AVG$\uparrow$   & LPIPS$\downarrow$ & Peak GPU Memory $\downarrow$ \\
    \midrule
    \midrule
        \usym{2718}  &   \usym{2718}    &   67.1    &  0.162     &37.9\\
        \usym{2718}  &    \usym{2714}   &   64.4    &  0.143     &37.9\\
        \usym{2714}  &   \usym{2718}    &   \textbf{67.3}    &   0.159 & \textbf{13.1}\\
        \usym{2714}  &   \usym{2714}    &   65.0    &   \textbf{0.138}    &   \textbf{13.1} \\
    \bottomrule
    \end{tabular}
    }
  \label{tab:table4}
\end{table}

\subsection{Ablation Study}

\paragraph{1) Effect of IDM and LFCM:}
Tab.~\ref{tab:table4} quantifies the individual contributions of our proposed modules. IDM primarily mitigates computational bottlenecks, slashing peak GPU memory usage from 37.9 GB to just 13.1 GB. Remarkably, it does so while slightly boosting AVG from 67.1\% to 67.3\%, confirming that the invertible design enables scalable deep diffusion optimization without sacrificing transferability. In contrast, LFCM ensures superior visual imperceptibility by suppressing unstable high-frequency perturbations, lowering LPIPS from 0.162 to 0.143. When combined, the full IDATA method achieves the best visual imperceptibility (LPIPS: 0.138) while maintaining low memory footprint (13.1 GB) and strong transferability (AVG: 65.0\%). This demonstrates that IDM and LFCM work synergistically to strike an optimal balance between attack success rate, perceptual quality, and computational efficiency.

\textit{2) Perturbation Timestep $\tau$:}
Fig.~\ref{pic:fig1} presents the ablation study on the perturbation timestep $\tau$, compared with state-of-the-art diffusion-based attacks.
To ensure fair comparisons, we enforce a strict perceptual budget $LPIPS \le 0.14$.
As shown, smaller $\tau$ leads to suboptimal transferability, while $\tau=24$ exceeds the budget.
Thus, we select $\tau=20$ as the default, achieving the highest transferability with a compliant $\text{LPIPS}=0.13$.

\begin{figure}[t]
    \centering
    \includegraphics[width=\linewidth]{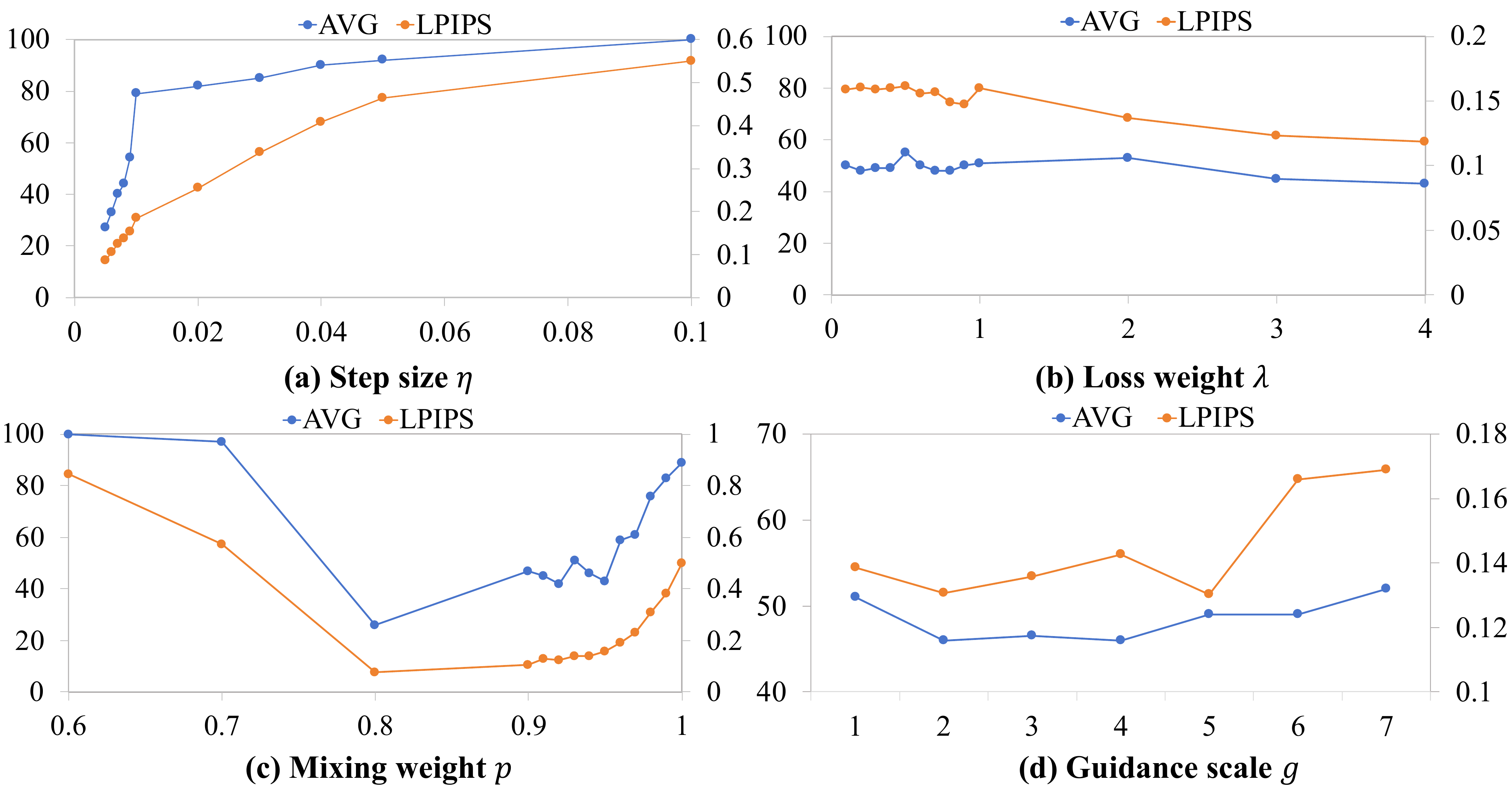}
    \caption{A quantitative study on parameter settings: step size $\eta$, loss weight $\lambda$, mixing weight $\textit{p}$ and guidance scale $\textit{g}$. The left y-axis shows AVG, and the right y-axis shows LPIPS.}
    \label{pic:ablations}
\end{figure}

\textit{3) Parameter Sensitivity: }
Fig.~\ref{pic:ablations} analyzes four key hyper-parameters: the step size $\eta$, the perceptual regularization weight $\lambda$, the mixing weight $p$, and the guidance scale $g$. 
In Fig.~\ref{pic:ablations}(a), 
increasing $\eta$ improves AVG but also elevates LPIPS. When $\eta > 0.01$, transferability gains become marginal while visual distortion accelerates, so we set $\eta=0.01$.
In Fig.~\ref{pic:ablations}(b), LPIPS shows an overall decreasing trend as $\lambda$ increases, while AVG reaches its peak at $\lambda=2$, indicating that a moderate perceptual regularization strength yields the best transferability-imperceptibility trade-off. 
Fig.~\ref{pic:ablations}(c) shows that larger $p$ generally improves visual imperceptibility but overly large values noticeably reduce AVG; the best balance is achieved at $p=0.93$.
As shown in Fig.~\ref{pic:ablations}(d), AVG remains relatively stable across different guidance scales, whereas LPIPS generally increases as $g$ becomes larger. 
Accordingly, we set $g=1$. 
Overall, IDATA remains reasonably robust under moderate hyper-parameter variations.

\textit{4) Frequency Analysis:}
Fig.~\ref{pic:ablation_vis} compares adversarial examples generated under different frequency constraints. Split-view visualization shows that all-frequency and high-frequency perturbations produce fine-grained, noise-like artifacts that impair visual quality. Conversely, low-frequency perturbations yield smoother, structure-aware patterns that preserve object textures and structural integrity, leading to substantially lower LPIPS scores. This validates the core motivation of LFCM: confining perturbations to low-frequency latent components inherently achieves a superior balance between attack imperceptibility and transferability.

\section{Conclusion}
In this paper, we propose \textbf{IDATA}, a scalable and memory-efficient invertible diffusion framework for unrestricted transferable adversarial attack. By integrating an Invertible Diffusion Module (IDM) with a Low-Frequency Constraint Module (LFCM), IDATA enables deep adversarial optimization with substantially reduced memory overhead, while achieving strong cross-model transferability and improved visual imperceptibility. Extensive experiments across diverse model architectures, defense settings, and datasets demonstrate that IDATA attains a favorable trade-off among attack effectiveness, computational efficiency, and perceptual quality, making it a promising tool for black-box robustness evaluation in real-world multimedia systems.


\begin{acks}
  This work was supported by the National Natural Science Foundation of China No. xxxxxxxx and No. xxxxxxxx.
\end{acks}

\bibliographystyle{ACM-Reference-Format}
\balance
\bibliography{sample-base}
\end{document}